**TITLE**

# Representation-guided in-context learning for medical image interpretation with multimodal large language models

**AUTHORS**

Minda Zhao[1†], Fangyu Hu[2, 3†], Yan Luo[1†], Yutong Yang[1], Jiahui Cai[1], Kaichen Zhou[1, 4], Manling Li[5], Paul Liang[4], Yilun Du[6, 7], Lucy Q. Shen[2*], Mengyu Wang[1, 6, 8, 9*]

[†]MZ, FH, and YL are co-first authors.

*LQS and MW are co-senior authors.

**AFFILIATIONS**

1. Harvard AI and Robotics Lab, Schepens Eye Research Institute of Massachusetts Eye and Ear, Harvard Medical School, Boston, MA, USA

2. Department of Ophthalmology, Massachusetts Eye and Ear, Harvard Medical School, Boston, MA, USA

3. Department of Ophthalmology, National Taiwan University Hospital, Taipei City, Taiwan

4. MIT Media Lab, Massachusetts Institute of Technology, Cambridge, MA, USA

5. Department of Computer Science, McCormick School of Engineering, Northwestern University, Evanston, IL, USA

6. Kempner Institute for the Study of Natural and Artificial Intelligence, Harvard University, Boston, MA, USA

7. Department of Computer Science, John A. Paulson School of Engineering and Applied Sciences, Harvard University, Boston, MA, USA

8. Harvard Data Science Initiative, Harvard University, Boston, MA, USA

9. Broad Institute of MIT and Harvard, Boston, MA, USA

**CORRESPONDING AUTHOR**

Mengyu Wang, PhD

Schepens Eye Research Institute

20 Staniford Street

Boston, MA 02114, USA

Email: mengyu_wang@meei.harvard.edu

**FINANCIAL SUPPORT**

LQS is supported by NEI R01EY031696. MW is supported by NIH grants R01 EY038687, R01 EY036222, and P30 EY003790.

**CONFLICT OF INTEREST**

LQS is a consultant for FireCyte Therapeutics, AbbVie Inc and Merck & Co, Inc.

## Abstract

Medical image interpretation is central to diagnosis and care, yet adapting general-purpose multimodal large language models (MLLMs) often requires resource-intensive domain-specific fine-tuning. Here we introduce representation-guided in-context learning (RG-ICL), a training-free inference framework that retrieves query-aligned demonstrations using frozen encoders, without task-specific parameter updates. Across eight datasets spanning histopathology, radiology and retinal fundoscopy, RG-ICL improved classification (mean gain 20 percentage points) and visual question answering (VQA) (mean gain 13 percentage points) over no-context and conventional ICL, approaching or exceeding training-based comparators. Which cases were retrieved mattered more than how many: 6 query-aligned cases outperformed up to 32 randomly selected ones, whereas fixed or random cases often reduced accuracy below baseline. For VQA, aligning reference cases with both image content and question intent produced further gains. These findings indicate that for medical image interpretation, curating which reference cases an MLLM sees is a practical alternative to retraining it.

# Introduction

Multimodal large language models (MLLMs), a class of foundation models capable of processing multiple input modalities including text and images, have emerged as broadly accessible artificial intelligence (AI) systems[1]. Publicly available general-purpose models, such as ChatGPT (OpenAI), Gemma (Google), and Qwen (Alibaba), are now widely used across diverse applications ranging from specialized tasks to everyday decision-making[2]. In medicine, where imaging plays an essential role in disease diagnosis and clinical care but also imposes substantial demands on clinical experts[3], MLLMs have also prompted growing interest as potential assistants for image interpretation, radiology report generation, and research support[4–8]. Early studies across radiology[9–11], pathology[12], and ophthalmology[13–16] suggest that general-purpose MLLMs can interpret medical images, answer visual questions, and integrate multimodal information for clinically oriented reasoning[8,17,18]. However, their performance generally remains inferior to conventional supervised fine-tuning approaches and is insufficient for real-world clinical deployment[5,10,19–21], raising critical safety concerns as inaccurate responses may lead to misdiagnosis and delayed treatment[7,22,23]. Although effective[4], conventional fine-tuning of MLLMs typically requires large annotated datasets, specialized technical expertise and considerable computational resources owing to the models' large parameter counts, limiting its feasibility in resource-constrained healthcare settings.[24]. Moreover, fine-tuned model performance can degrade sharply outside benchmark conditions, limiting their applicability to the narrow, task-specific datasets on which models are trained[25] and restricting adaptability across imaging modalities, patient populations, and diverse clinical tasks[6,8].

Recent work has highlighted the potential of in-context learning to enhance frozen (retraining-free, without any parameter updates) MLLM performance while preserving adaptability across clinical tasks[25]. By providing task-specific prompts and a small number of labeled reference images at inference time, in-context learning enables task adaptation without model retraining[26]. Recent general-domain multimodal ICL studies have developed retrieval strategies based on visual, textual, or task-aware representations, but these approaches have not been systematically evaluated across diverse medical imaging tasks[27–30]. Several studies have reported MLLM performance improvements using in-context learning in select medical image classification tasks in pathology[31,32], ophthalmology[33], and radiology[10,34]. However, reported gains have been variable and validated only in narrow settings using a small number of datasets, while concerns regarding the reliability persist. Importantly, visual in-context learning is highly sensitive to demonstration selection[35]; How to select context samples that consistently yield substantial performance gains across medical imaging tasks remains unresolved.

We hypothesized that medical in-context learning would benefit from selecting reference cases that are closely aligned with the query in latent representation space. Medical images contain diverse forms of clinically relevant visual information, such as anatomical structures and pathological findings, that may be jointly encoded within a unified latent representation rather than prespecified as distinct retrieval dimensions. For visual question answering (VQA), reference-case relevance depends on both visual similarity and question alignment, as visually similar images may be paired with questions targeting different anatomical structures or clinical findings. Drawing on advances in retrieval-augmented generation[36], transferable vision-language representations[37], and self-supervised visual representations[38], we reasoned that reference cases aligned with the query in representation space could provide more relevant context, thereby improving medical image inference without model retraining.

To test this hypothesis, we developed representation-guided in-context learning (RG-ICL), a training-free framework that retrieves, for each query, the nearest reference cases in representation spaces defined by frozen vision and language encoders and presents them as context to a frozen MLLM. RG-ICL leaves the MLLM, the prompt template, the decoding settings, and the context size fixed; the only adaptive component is context sample selection. Whereas prior medical ICL studies have often focused on narrower modality or disease settings, or used domain-specific retrievers, RG-ICL evaluates frozen, general-purpose vision and language encoders across heterogeneous medical imaging tasks. CLIP and DINOv3 have shown promise in medical vision without domain-specific pretraining. To the best of our knowledge, their use for MLLM demonstration retrieval has not been systematically evaluated across heterogeneous medical modalities, task formats and model families.

We evaluated RG-ICL across eight public medical vision-language datasets spanning disease classification and open-ended VQA (Fig. 1a), comparing it against no-context prompting, fixed and random context controls (Fig. 1b-d), and training-based approaches, including two supervised image classifiers (ResNet-50 and ViT) and a parameter-efficient supervised fine-tuned MLLM (Gemma4 LoRA). Across modalities, task formats and model families, we examined three questions. (i) Does RG-ICL outperform no-context prompting, conventional ICL and training-based baselines? (ii) Does performance depend more on demonstration identity than quantity, and does joint image-question alignment improve VQA? (iii) Do these effects extend across general-purpose, medically oriented and compact MLLMs, and when does retrieved context improve or harm predictions? Together, these analyses establish RG-ICL as a broadly evaluated, training-free strategy for medical multimodal inference across heterogeneous tasks and model settings.

# Results

## Representation-guided context sample retrieval framework for medical vision-language in-context learning

We define RG-ICL as a training-free inference framework in that it requires no task-specific parameter updates to the MLLM, representation encoders, or retrieval model. It instead retrieves query-aligned demonstrations from an existing labeled reference pool and presents them to a frozen MLLM (Fig. 1d, RG-ICL). For each dataset, the training set served as the candidate pool for context sample selection (Fig. 1e), while the testing set was used exclusively as queries (Fig. 1b). Each context sample paired an image with corresponding label (for classification) or a question-answer pair (for VQA) (Fig. 1c). For each held-out query, RG-ICL ranked candidate reference cases by representation-based similarity and inserted six highest-ranked cases into the dataset-specific multimodal prompt (Fig. 1f, Supplementary Fig. 1 and 2), from which the MLLM generated either a structured diagnostic prediction (for classification tasks) or a free-form response (for VQA tasks). Representations were computed using frozen general-purpose encoders: CLIP (Contrastive Language-Image Pre-training) for whole-image semantic alignment, DINOv3 (self-DIstillation with NO labels version 3) for fine-grained visual similarity, and BGE (BAAI General Embedding) for question-intent alignment in VQA. A fused CLIP+DINOv3 approach with equal weighting and a BGE+DINOv3 approach weighted 0.3:0.7 were also evaluated for combined semantic-morphological and visual-textual alignment, respectively.

To isolate the contribution of query-aligned context sample selection from context quantity, RG-ICL was against three controls (Fig. 1d). The zero-shot (No-ICL) condition measured baseline MLLM performance from the query alone. The fixed-context (Fixed-6) condition provided the same six randomly selected context samples to every query, controlling for prompt length, task demonstration and label-space exposure. The random-context (Random-6) condition provided six context samples drawn randomly per query, controlling for in-domain example exposure without query-specific selection. The RG-ICL condition used the same six-case budget but selected references by representation alignment with the query. Across all in-context comparisons, the MLLM, representation encoders, prompt template, and context size were held fixed, isolating context sample selection as the experimental variable.

We evaluated RG-ICL on eight established public medical vision-language datasets spanning two major modes of medical MLLM inference: closed-form disease classification and open-ended visual question answering (VQA) (Fig. 1a). The classification suite comprised breast histopathology (BreakHis[39]), chest radiography (TBX11K[40]), diabetic retinopathy severity grading (DDR[41]) and glaucoma assessment from retinal fundoscopy (LAG[42]), whereas the VQA suite extended the evaluation to heterogeneous medical figures, pathology and radiology (SLAKE[43], PathVQA[44], VQA-RAD[45] and VQA-Med2019[46]). Together, these datasets evaluate whether RG-ICL generalizes across imaging modalities, organ systems, and task formats, supporting both diagnostic decisions and image-question reasoning. We further evaluated RG-ICL across current frontier MLLM families, including open general-purpose models (Qwen3.6[47] and Gemma4[48]), a medically oriented open model (MedGemma[49]) and GPT-5.5[50] as a closed-source comparator, and compared these training-free results with supervised ResNet[51] and ViT[52] image-classification models and Gemma4 LoRA[53] supervised fine-tuning as training-based adaptation comparators. Classification performance was measured using accuracy and area under the receiver operating characteristic curve (AUC) from parsed model outputs, and VQA responses were evaluated with a prespecified GPT-5.4-mini judge[4,54–56] across semantic accuracy, completeness, factuality and conciseness (Supplementary Fig. 3).

**Representation-guided in-context learning improves medical image classification across modalities and tasks**

Providing frozen MLLMs with visually aligned reference cases at inference improved medical image classification across all four datasets and evaluated MLLM backbones, approaching the performance of task-specific supervised models. Relative to No-ICL, RG-ICL increased diagnostic accuracy by a mean absolute difference of approximately 0.20, with gains reaching 0.40 for MedGemma on TBX11K (Fig. 2). The largest mean absolute gains were observed on BreakHis and TBX11K, at 0.24 and 0.30, respectively, whereas the gain on DDR was more modest, at 0.046. Performance varied across CLIP, DINOv3 and CLIP+DINOv3 retrieval, with maximum differences of 0.13 in accuracy and 0.10 in AUC; CLIP+DINOv3 most frequently achieved the strongest performance. Eight-subtype histopathology classification on BreakHis showed a similar pattern, with CLIP+DINOv3 achieving the highest accuracy (Extended Data Fig. 1).

In contrast, conventional ICL with fixed or randomly selected context samples (Fixed-6 and Random-6) provided no comparable benefit and frequently degraded performance below No-ICL (Fig. 2). For example, Qwen3.6’s accuracy on TBX11K fell from 0.602 with no context to 0.467 when given fixed context and to 0.448 when given randomly selected context. This indicates that the gains from RG-ICL are associated with visually relevant retrieved examples rather than context length alone. Notably, this training-free improvement is competitive with, and in some cases exceeds, resource-intensive supervised models: Gemma4 with CLIP+DINOv3 surpasses the supervised ResNet50 on BreakHis (0.871 versus 0.866) and outperforms fine-tuned Gemma4 LoRA on LAG (0.937 versus 0.930), while on TBX11K it's only less than 3 percentage points behind ResNet50, ViT224, and Gemma4 (0.964 versus 0.991, 0.989, 0.980, respectively), all without any gradient updates or task-specific fine-tuning. The benefit of RG-ICL on BreakHis was also consistent across microscopic magnifications (40x, 100x, 200x, and 400x), suggesting robustness to variation in tissue scale and morphological detail (Supplementary Fig. 4).

Representative cases illustrate the benefit of RG-ICL at the individual image level across the four classification datasets (Fig. 4a). In each example, No-ICL, Fixed-6, and Random-6 ICL approaches produced incorrect answers, whereas RG-ICL with CLIP+DINOv3 directed the model to the correct label by supplying six context samples matched to the query in imaging modality, global appearance, and morphological features. In the example of BreakHis, CLIP+DINOv3 retrieved reference images with similar tissue architecture and cellular morphology with the query. In the case of TBX11K, retrieved chest radiographs were matched in projection view and overall cardiothoracic appearance. These cases reflect the findings above: visually aligned context sample selection improves diagnostic accuracy, while visually irrelevant context provides little benefits and can degrade model performance.

Further analyses using Gemma4-31B on LAG with 10 different vision encoders showed that RG-ICL remained effective across encoder families, while CLIP+DINOv3 achieved the highest accuracy and AUC (Extended Data Fig. 2). Applying RG-ICL on top of Gemma4 LoRA checkpoints for TBX11K and LAG yielded additional accuracy and AUC gains over LoRA alone, supporting a complementary rather than competing relationship between RG-ICL and supervised fine-tuning (Extended Data Fig. 3). Replacing Gemma4-31B and MedGemma-27B with the compact versions (Gemma4-E4B and MedGemma-4B) on LAG showed that RG-ICL remains effective with compact MLLMs (Extended Data Fig. 4). We also examined the effect of context sample order by comparing most-similar-first, most-similar-last, and random orderings on LAG and BreakHis, which revealed modest yet dataset-dependent sensitivity to the context sample demonstration order (Extended Data Fig. 6).

**Representation-guided in-context learning improves medical visual question answering across modalities, with the addition of text-similarity retrieval generally producing the strongest gains**

Across the SLAKE, PathVQA, and VQA-Med2019 datasets and various MLLM backbones, BGE+DinoV3 generally achieved the highest semantic accuracy, completeness, and factuality scores (Fig. 3). This approach not only outperformed No-ICL and vision-only RG-ICL methods but also surpassed the supervised fine-tuned baseline (Gemma4 LoRA SFT) in semantic accuracy and completeness for every backbone tested, and in factuality for all but two cases, with gains as large as 0.662 versus 0.444 (GPT-5.5, VQA-Med2019, semantic accuracy). An exception was observed in the VQA-RAD dataset, but only for the joint textual and visual retrieval approach: BGE+DinoV3 slightly underperformed No-ICL across all four MLLM backbones (e.g., Gemma4 factuality dropped form 0.712 with No-ICL to 0.685 with BGE+DinoV3), whereas the vision-only retrieval methods (CLIP, DinoV3, and CLIP+DinoV3) continued to deliver consistent gains and remained competitive with or exceeded Gemma4 LoRA SFT (Gemma4 CLIP+DinoV3 factuality of 0.745 versus

Gemma4 LoRA's 0.683). Of note, the consistent benefit of BGE+DINOv3 and the exception observed on VQA-RAD were also seen under traditional lexical metrics (Exact Match, BLEU-4, ROUGE-L, and METEOR), supporting the robustness of the findings (Supplementary Fig. 5).

Consistent with the image classification tasks, the conventional ICL methods (Fixed-6 and Random-6) offered minimal benefit and frequently impaired model performance in semantic accuracy, completeness, and factuality. Using SLAKE as an example, Fixed-6 and Random-6 reduced Qwen3.6's performance from a 0.696 baseline to 0.413 and 0.477 in semantic accuracy, from 0.706 to 0.413 and 0.478 in completeness, and from 0.759 to 0.476 and 0.548 in factuality, respectively. Interestingly, conciseness scores remained uniformly high (mostly above 0.90) across nearly all datasets, MLLMs, and retrieval methods, suggesting that these MLLMs might be inherently good at adhering to length constraints regardless of the context provided.

Representative VQA cases further illustrate the benefit of joint visual and question-intent alignment at the individual query level across the four datasets (Fig. 4b). In each example, No-ICL, Fixed-6, and Random-6 produced inaccurate answers, whereas RG-ICL with BGE+DINOv3 directed the model towards responses with higher semantic accuracy, completeness, and factuality by supplying six context samples aligned with the query in both visual content and question intent. For instance, in the example case of SLAKE, BGE+DINOv3 retrieved examples that were all CT images of similar anatomical sections, with questions consistently addressing the spleen, matching the query at both the visual and semantic level. In the VQA-RAD example, six axial brain sections from similar anatomical levels were retrieved, with the associated questions mostly addressing acute cerebrovascular findings, reflecting coherent alignment with both the visual appearance and the clinical question being asked. These suggest that the semantic alignment of the clinical question is a meaningful driver of RG-ICL benefit beyond visual similarity alone. Complete instantiated prompts and raw model outputs for representative LAG classification and VQA-RAD cases, in which RG-ICL corrected errors made by No-ICL, Fixed-6 and Random-6 ICL, are provided in Supplementary Fig. 6.

**Retrieval relevance outweighs additional non-targeted context**

The failure of Fixed-6 and Random-6 to produce comparable gains suggests that the benefit of RG-ICL may stem from visual relevance of retrieved references, rather than from the presence of additional examples alone. To test this interpretation directly, we performed two complementary analyses on the classification tasks.

First, in a random-context analysis, we quantified for each query the mean CLIP cosine similarity between the query image and its randomly sampled reference cases, and examined whether this incidental visual similarity predicted query-level correctness (Fig. 5a). Higher random-context similarity was associated with a greater probability of correct classification in multiple dataset-model combinations. For example, on TBX11K, fitted correctness increased from approximately 0.27-0.68 in the lowest similarity range to approximately 0.61-0.83 in the highest similarity range across MLLMs. Similar positive trends were observed on BreakHis, where Gemma4 increased from approximately 0.48 to 0.79, and on LAG, where Gemma4 and MedGemma increased from approximately 0.63 to 0.80 and from approximately 0.49 to 0.78, respectively. Logistic regression confirmed significant positive associations between visual similarity and per-query correctness across multiple dataset-model combinations ($P <0.05$), indicating that even when examples were sampled at random, visually aligned contexts were more likely to support correct predictions.

The context-count analysis was consistent with this interpretation (Fig. 5b). On LAG with Gemma4, increasing the number of fixed or randomly selected context samples from k = 2 to k = 32 did not close the performance gap to visually retrieved context: fixed-context accuracy ranged from approximately 0.62 to 0.82 and random-context accuracy from approximately 0.68 to 0.77, whereas CLIP+DINOv3 retrieval achieved approximately 0.93 with only six retrieved examples. A similar pattern was observed for AUC, with fixed-context and random-context settings remaining at approximately 0.77-0.83 and 0.70-0.76, respectively, compared with approximately 0.97 for CLIP+DINOv3 retrieval. On LAG with Gemma4, six visually aligned demonstrations therefore outperformed up to 32 fixed or randomly selected demonstrations. Within this dataset, model and tested K range, query alignment contributed more than increasing non-targeted context alone.

**Medical VQA requires joint alignment of visual content and question intent**

Medical VQA extends beyond image classification by requiring the MLLM to process an image-question pair jointly. A visually similar reference image may provide misleading context if its associated question addresses a different finding. Effective retrieval for VQA should therefore select references that are aligned with both the visual content of the query image and the semantic intent of the clinical question. We examined this through two complementary analyses: a changed-answer analysis across all RG-ICL conditions, and a question-aware retrieval analysis on SLAKE.

In the changed-answer analysis, we pooled all RG-ICL settings and identified queries for which the ICL-induced answer differed from the No-ICL baseline. Each such change was labelled as a correction (incorrect → correct) or a harm (correct → incorrect) using the LLM-judge perfect-score endpoint. Retrieved reference questions were then ranked by BGE cosine similarity to the query question, and the probability that a changed answer was a correction was modelled with logistic regression. Across dataset-model combinations, the probability of correction increased with retrieved-question similarity (Fig. 6a), with statistically significant positive associations in multiple settings ($p < 0.05$), indicating that ICL-induced answer changes are more likely to be beneficial when the semantic content of retrieved questions resembles that of the query.

To test this principle directly, we evaluated BGE+DINOv3 fusion retrieval on SLAKE, combining question-intent alignment via BGE text embeddings with visual similarity via DINOv3. BGE+DINOv3 improved semantic-perfect accuracy above the CLIP+DINOv3 visual-only baseline across all MLLM backbones, with the configuration that weighted DINOv3 visual similarity more heavily than BGE question similarity (0.7 visual, 0.3 textual) providing the most consistent gains (Fig. 6b). Together, these analyses suggest that effective context sample retrieval for medical VQA should align reference cases with both the image findings and the clinical question being asked, and that incorporating question-intent alignment can improve upon visual retrieval without sacrificing the underlying visual signal.

## Discussion

This study demonstrates that representation-guided in-context learning can improve general-purpose MLLMs for medical vision-language inference with a small set of visually-aligned reference cases, without updating model parameters. Across four public classification benchmarks spanning histopathology, chest radiography and retinal fundus imaging, RG-ICL outperformed no-context, fixed-context and random-context prompting across multiple MLLM families, yielding a mean absolute accuracy gain of approximately 0.20 over No-ICL while approaching or exceeding task-specific supervised models in several settings. Similar improvements were observed in the four medical VQA datasets, with additional incorporation of question intent alignment alongside visual similarity yielding the highest gains.

Our results contrast with prior studies, which reported suboptimal diagnostic performance of general-purpose MLLMs for medical image interpretation[9–16], as well as limited and variable gains from naive in-context learning[33], relative to conventional training-based approaches. This may be partly clarified by the fixed-context and random-context controls in this study, where providing random, visually irrelevant references did not provide benefit and often degraded performance. The random-context similarity and context-count ablation analyses in Fig. 5 further showed that the benefit of RG-ICL does not stem from the quantity of context samples, but their resemblance and relevance to the query. These findings support a case-based interpretation of RG-ICL, in which reference cases selected for their proximity to the query in learned representation spaces support more accurate diagnostic predictions. The relative performance of CLIP, DINOv3 and CLIP+DINOv3 supports this view. CLIP captures global, language-aligned image content, whereas DINOv3 contributes complementary sensitivity to fine-grained structural and textural variation; combining the two most frequently yielded the strongest classification performance. Comparisons across common visual representation models with different encoding mechanisms on LAG also favored the fused CLIP+DINOv3 approach (Extended Data Fig. 2). The VQA results further extend this principle from visual similarity to multimodal clinical reasoning. By combining BGE question-intent similarity with DINOv3 visual similarity, it generally achieved stronger performance than visual-only retrieval for VQA, where the image and question input are both crucial for inference. Notably, the VQA-RAD exception, where BGE+DINOv3 slightly underperformed no-context prompting while vision-only RG-ICL approaches remained beneficial, suggests that generic text similarity may sometimes overweight superficial wording, and that such approach should be clinically grounded and tailored by case. Future VQA retrieval could distinguish different dimensions of question content, such as anatomical target, imaging modalities, and diagnostic tasks, rather than relying on global similarity alone.

RG-ICL has practical implications for clinical deployment in settings where model retraining is expensive, technically challenging, or constrained by data governance. In contrast to prior approaches that rely on retraining or fine-tuning foundation models or deploying domain-specific vision encoders, RG-ICL operates entirely with frozen, general-purpose MLLMs and representation models, with prompt-level reference cases being the only adaptive component. RG-ICL avoids the need to fine-tune the MLLM or train a task-specific retriever, but it does not eliminate labeled-data requirements. Instead, it reuses an existing labeled reference pool and selects a small query-specific subset for each prompt. Its practical utility will therefore depend on the size, coverage and quality of the available reference pool. Furthermore, it can be deployed on top of any frontier MLLM, including both open and closed-source models accessed through an application programming interface (API). A hospital or laboratory could update the reference pool as clinical cases accumulate, allowing adaptation to local imaging machines, imaging protocols, patient populations and disease prevalence without changing model weights. RG-ICL is also complementary to conventional supervised adaptation (Extended Data Fig. 3). While supervised fine-tuning encodes task-specific decision boundaries, RG-ICL can supply instance-level evidence at inference time. On LAG, RG-ICL remained effective with compact Gemma4-E4B and MedGemma-4B models and across biomedical and general-purpose retrieval encoders; feature-space analyses across all eight datasets further revealed strong dataset-level organization but dataset-dependent label separation (Extended Data Figs. 2, 4 and 5). Performance was only modestly sensitive to demonstration order, and an Adaptive-K controller nearly preserved fixed K = 6 performance with a lower mean terminal demonstration budget (Extended Data Figs. 6 and 7). Together, this study demonstrates that by optimizing context sample selection based on representation-level similarity, RG-ICL has the potential to become a more reliable, accessible, and adaptable AI framework for medical image interpretation in clinical practice.

Several limitations of this study warrant consideration. The study evaluates public benchmark datasets rather than prospective clinical workflows, which may under-represent demographic heterogeneity, scanner variation, rare or borderline cases, and real-world artefacts. The VQA evaluation relied on an LLM judge, which enables scalable assessment but does not replace expert clinical evaluation. Future work should be complemented by blinded clinician review. Finally, the experiments used English prompts and predominantly English VQA annotations; generalizability to multilingual clinical workflows and institution-specific terminology remains to be evaluated. Prospective validation across healthcare settings, image platforms, and languages will be necessary before clinical deployment.

# Methods

We performed a comprehensive evaluation of representation-guided in-context learning (RG-ICL) across medical image classification and visual question answering (VQA). For each dataset, the training set was used as the candidate pool for in-context reference selection, and held-out test examples were used only for final evaluation. Test examples were excluded from all retrieval pools. RG-ICL was evaluated as a training-free inference procedure: the MLLMs, representation encoders and retrieval backbones remained frozen, with no task-specific parameter updates or retriever training. The procedure nevertheless required labeled reference cases from the training partition. In the principal comparisons, fixed-context, random-context and representation-guided conditions all used the same six-reference budget (K=6).

**Datasets and preprocessing**

**Dataset selection and scope.** We evaluated RG-ICL on eight public medical datasets selected to span two major task formats: closed-form diagnostic classification and open-ended visual question answering (VQA). The classification suite comprised BreakHis[39], which tests benign-malignant breast histopathology classification across 40X, 100X, 200X and 400X magnifications; TBX11K[40], which tests chest radiograph classification across healthy, sick but non-TB and TB labels; DDR-512[41], which tests diabetic retinopathy (DR) classification across six categories (no DR, mild non-proliferative DR, moderate non-proliferative DR, severe non-proliferative DR, proliferative DR and ungradable) from retinal fundus photographs; and LAG[42], which tests glaucoma assessment from retinal fundus structure. The VQA suite comprised SLAKE[43] for heterogeneous medical visual question answering, PathVQA[44] for pathology question answering, and VQA-RAD[45] and VQA-Med2019[46] for radiology question answering. Together, these datasets evaluate RG-ICL across histopathology, chest radiography, retinal fundus imaging, general medical figures, pathology and radiology, and across both discrete diagnostic decisions and image-question reasoning.

**Data partitioning and retrieval pools.** Across all datasets, retrieval pools were restricted to training or reference partitions, and held-out test examples were excluded from retrieval. Official training, validation and test partitions were used when public labels were available. For BreakHis and LAG, which do not provide official splits matching our evaluation protocol, we generated train/validation/test partitions using a 70/15/15 ratio with seed 3407. The BreakHis split was performed at the patient level, and magnification-specific evaluations were retained for supplementary analyses. For BreakHis, benign versus malignant tumor classification was used as the primary evaluation in the main analysis, with eight-subtype histopathology classification reported as an extended analysis. For TBX11K, because the official test labels are not publicly released, we used the official validation set as the held-out test set and split the official training set into training and validation subsets using an 85/15 ratio with seed 3407.

### Model inference, prompt construction and context conditions

**MLLM inference.** We evaluated RG-ICL using two current high-capacity open-source MLLMs, Qwen3.6-27B[47] and Gemma4-31B[48], together with a medically oriented open MLLM, MedGemma-27B[49], and a closed-source frontier comparator, GPT-5.5[50]. This model panel was selected to cover contemporary open-source, closed-source and medical-domain MLLM settings under a shared evaluation protocol. Following recent medical LLM benchmarking and medical question answering evaluation settings[57,58], inference used temperature 1.0 as a standardized sampling configuration to balance deterministic and diverse generation; this setting was fixed across all models, datasets and context conditions rather than tuned for individual tasks. Thinking modes were disabled across all models.

**Prompt construction.** Prompts used dataset-specific task instructions, label vocabularies and output contracts. For classification, binary tasks required a predicted label and a probability for the positive class, whereas multiclass tasks required a predicted label and a probability distribution over all classes; prompts also requested a compact confidence field and a short evidence field. When context samples were used for classification, labelled reference images were placed before the query image according to the assigned context condition. For VQA, prompts included the query image and natural-language question, with each context sample consisting of an image, its associated question and its answer. Dataset-specific VQA prompts requested concise short answers or compact structured answers according to the task contract. (See Supplementary Fig. 1 and 2 for details.)

## Representation-guided retrieval framework

**Retrieval formulation.** For each dataset, let the training set be the reference set $D_{ref}=\{r_i\}_{i=1}^{N}$, where each reference case contains an image and its supervision: a diagnostic label for classification, or an image-question-answer triplet for VQA. For a held-out query $u$, where $u=x$ for classification and $u=(x,q)$ for VQA, a context-selection function retrieves $K$ reference cases from the training set:

$$C_K(u)=TopK_{r_i \in D_{ref}}\, s(u,r_i).$$

The selected references are inserted into the dataset-specific multimodal prompt, and the fixed MLLM produces either a structured diagnostic prediction or a free-form VQA answer:

$$\hat{o} = f_\theta\left(P\left(u, C_K(u)\right)\right).$$

Here, $P(\cdot)$ is the prompt constructor, $s(u, r_i)$ is the retrieval score, $f_\theta$ is the fixed MLLM, and $\hat{o}$ is the parsed classification output or generated VQA response. The only adaptive step is the choice of $C_K(u)$ at inference time. In the main experiments, $K$ was fixed at 6 for fixed-context, random-context and representation-guided ICL, matching the six-shot setting used throughout the principal comparisons; $K$ was varied only in the context-count, fixed-budget scaling and Adaptive-K analyses.

**Context conditions.** Having defined RG-ICL as the selection of $C_K(u)$, we constructed a set of context controls to isolate what aspect of in-context learning drives performance. The no-context condition measured the baseline ability of each MLLM from the query image, or the query image-question pair, alone. The fixed-context condition added six reference cases but kept them identical for every query, controlling for prompt length, label exposure and the presence of worked examples. The random-context condition also used six reference cases but sampled them from the training set for each query without using query similarity, controlling for exposure to labelled cases from the same dataset without targeted selection. The RG-ICL condition used the same six-case budget but selected the references by representation similarity to the query. Thus, across the principal in-context comparisons, the model, prompt template, decoding setting and number of examples were held fixed; the experimental variable was whether the reference cases were generic, randomly sampled or query-aligned.

**Representation roles.** In RG-ICL, reference cases are selected not simply as additional examples, but as case-based scaffolds for inference. Reference cases that closely match the query in both overall image content and fine-grained visual structure are more likely to support accurate medical image interpretation. For VQA, their relevance may additionally depend on semantic alignment with the query question. We therefore operationalized reference-case relevance along four representation-based dimensions using frozen encoders. The first dimension is whole-image semantic alignment, we operationalized this dimension with Contrastive Language-Image Pre-training (CLIP)[37], which learns language-aligned visual representations through image–text contrastive learning. CLIP-based retrieval ranked candidate cases according to cosine similarity between whole-image embeddings, thereby testing whether overall, language-aligned image resemblance identifies useful in-context references. The second dimension was fine-grained visual similarity. We operationalized it with DINOv3[38], a self-supervised vision transformer that learns transferable visual representations without task-specific labels. DINOv3 representations retain fine-grained structural, textural and spatial information that may jointly reflect anatomical and pathological findings. DINOv3-based retrieval therefore tested whether close visual correspondence at this level provides an effective signal for reference selection. The third dimension was combined visual similarity. We operationalized this dimension by fusing the normalized CLIP and DINOv3 cosine-similarity scores. This condition tested whether combining whole-image, language-aligned similarity with fine-grained visual similarity improves the identification of useful reference cases. The fourth dimension, evaluated for VQA, was question-intent alignment. Visually similar images may be paired with questions targeting different anatomical structures, pathological findings or clinical tasks. We therefore used BGE[59], a text-embedding model that places semantically similar questions near one another in representation space, to quantify similarity between query and reference questions, and combined this score with DINOv3-based visual similarity. This condition tested whether jointly matching reference cases to the query image and question improves retrieval for open-ended medical VQA.

**Retrieval scores.** All embeddings were L2-normalized and compared using cosine similarity:

$$\cos(a,b)=\frac{a^{\top}b}{\|a\|_2\|b\|_2}.$$

For classification, the main fused visual retrieval score was

$$s_{vis}(x,x_i)=\alpha\, s_{CLIP}(x,x_i)+(1-\alpha)\, s_{DinoV3}(x,x_i),$$

where $s_{CLIP}$ measures whole-image semantic similarity and $s_{DinoV3}$ measures morphology-sensitive visual similarity. The principal CLIP+DINOv3 setting used $\alpha=0.5$.

For VQA, the main question-aware retrieval score was

$$s_{VQA}((x,q),(x_i,q_i))=\lambda\cos(h(q),h(q_i))+(1-\lambda)\, s_{DinoV3}(x,x_i),$$

where $h(q)$ is the BGE embedding of the query question. The principal BGE+ DINOv3 setting used $\lambda=0.3$, adding question-intent alignment while retaining DINOv3 visual similarity as the dominant retrieval signal.

**Retrieval index and leakage prevention.** For each held-out query, similarity scores were computed only against examples in the training split. Test examples were excluded from the retrieval index. Retrieved classification demonstrations comprised an image and its ground-truth label, whereas VQA demonstrations comprised an image, question, and reference answer.

**Fixed-budget scaling and Adaptive-K inference harness**

**Fixed-budget scaling.** To characterize performance as a function of demonstration budget, we evaluated Gemma4-31B at K ∈ {0, 2, 4, 6, 8, 16}. On LAG, K = 0 corresponded to No-ICL, and Random ICL and CLIP+DINOv3 RG-ICL were evaluated at each positive budget. On BreaKHis, CLIP+DINOv3 RG-ICL was evaluated at $K \in \{2, 4, 6, 8, 16\}$ separately for each magnification, with magnification-specific predictions pooled only for aggregate reporting. For each query and method, the reference cases at smaller budgets were exact prefixes of the $K = 16$ ordering. The train-only retrieval pools, prompt templates, decoding settings and seed 3407 were otherwise held fixed.

**Adaptive-K controller.** Adaptive-K was implemented as an online active-subset controller. For each query, the controller first performed No-ICL inference at $K = 0$ and stopped when the model-reported confidence was at least 0.95. Otherwise, it sequentially evaluated RG-ICL at $K = 2, 4, 6, 8$ and 16. At each positive budget below 16, inference stopped only when retrieval support was greater than 0.60 and the predicted label was unchanged from the immediately preceding budget. Retrieval support was defined as the fraction of the top-K reference labels matching the model's current prediction. Queries that did not satisfy both criteria advanced to the next budget and were forced to stop at $K = 16$. Performance was computed from each query's terminal prediction, and mean terminal K, cumulative prompt tokens, number of model calls and sequential latency were recorded. The same heuristic thresholds were applied to LAG and BreaKHis.

**Supervised and LoRA baselines**

**Supervised image baselines.** To contextualize training-free MLLM retrieval against conventional parametric adaptation, ResNet-50 and ViT-224 image classifiers were trained as supervised image-only comparators for the classification datasets. Models were optimized on the corresponding training splits using cross-entropy supervision and dataset-specific augmentation, selected by validation performance and evaluated only on the held-out test splits.

**Parametric MLLM adaptation via LoRA.** Low-rank adaptation (LoRA)[53] of the Gemma4-31B instruction-tuned backbone was used as the parametric MLLM adaptation comparator. Following recent parameter-efficient fine-tuning practice in medical LLM studies[60,61],LoRA adapters were applied to the language-decoder attention and MLP projection modules, with rank 16, alpha 32, dropout 0.05, learning rate $1 \times 10^{-4}$. Vision and projector modules were excluded for the classification LoRA runs. Best checkpoints were selected via validation performance, and final test inference utilized generation-based evaluation on the held-out splits.

**Complementarity analysis (LoRA plus retrieval).** To test whether inference-time retrieval complements parametric fine-tuning, the selected LoRA checkpoints were held fixed and used as base models for RG-ICL (Extended Data Fig. 3). CLIP+DINOv3 retrieval was then used to insert top-2, top-4 or top-6 reference examples at inference time where directly comparable runs were available. Reported deltas were computed relative to the corresponding LoRA-only checkpoint for the same dataset and metric, estimating the incremental contribution of instance-level retrieval beyond the fine-tuned parametric prior.

## Evaluation metrics, ablation analyses and statistical analysis

**Evaluation metrics.** Classification performance was quantified using accuracy and the area under the receiver operating characteristic curve (AUC), computed from parsed model outputs on the official evaluation splits of each dataset. For VQA dataset, generative responses were evaluated utilizing a pre-specified LLM-as-a-judge framework (GPT-5.4-mini). The evaluator was provided with the query, reference answer, and model prediction, and instructed to score the prediction from 0 to 100 across four dimensions: semantic accuracy (alignment with reference meaning), completeness (coverage of required information), factuality (absence of unsupported or contradictory claims), and conciseness (directness and brevity). The judge was constrained to leverage medical knowledge solely to recognize harmless synonyms, abbreviations, and equivalent phrasings (the full judge prompt is provided in the Supplementary Fig. 3). The primary VQA endpoint reported was the fraction of responses achieving a perfect score of 100 on each dimension.

**Ablation analyses.** Experimental ablations were designed to isolate the mechanisms driving ICL performance. To evaluate context quantity versus visual relevance, a context-count ablation analysis was conducted on the LAG dataset using Gemma4, sweeping the number of fixed- and random-context demonstrations against the principal CLIP+DINOv3 retrieval reference at K = 6. To validate the question-aware retrieval strategy, the SLAKE dataset was used to evaluate the BGE+DINOv3 score under varying modality weights, comparing configurations (including image-only retrieval and the principal 0.3:0.7 BGE:DINOv3 mixture) against the standard CLIP+DINOv3 baseline. For VQA answer-transition dynamics, ICL-induced changes were strictly dichotomized into corrections (incorrect-to-correct) or harms (correct-to-incorrect) based on the LLM-judge perfect-score endpoint. Finally, to assess complementarity with parametric adaptation, RG-ICL was layered atop the Gemma4 LoRA supervised fine-tuning checkpoints, with performance reported as percentage-point accuracy differences relative to the standalone LoRA baseline.

**Additional robustness and representation analyses.** On LAG, the compact-model analysis compared Gemma4-E4B and MedGemma-4B with their Gemma4-31B and MedGemma-27B counterparts while holding the split, retrieval pool, K = 6, prompt, decoding configuration, seed and frozen feature caches fixed. The retrieval-backbone analysis used Gemma4-31B at K = 6 and varied only the frozen representation used to rank reference cases. For the demonstration-order analysis, the same six retrieved cases were held fixed for each query and presented in descending similarity, ascending similarity or a seed-3407 random permutation. For feature-space visualization, normalized CLIP and DINOv3 image embeddings and BGE question embeddings were projected using UMAP with cosine distance, 30 neighbours, a minimum distance of 0.1 and seed 3407; 5-nearest-neighbour purity and silhouette were calculated in the original normalized feature space.

**Statistical analysis.** Logistic regression models were fitted via maximum-likelihood estimation on per-query outcomes to statistically evaluate the impact of visual relevance. For classification random-context trials, queries were grouped into deciles of mean CLIP similarity (between the query and randomly drawn references), and per-query correctness was modeled using mean similarity as the sole covariate. For VQA changed-answer analyses, the probability of an answer rescue was modeled as a function of the retrieved-question BGE-similarity percentile. In these regression models, significance was assessed at the level of the slope coefficient, with configurations yielding a raw $p < 0.05$ flagged as solid curves in the visualizations. All other comparative evaluations are reported as aggregated point estimates on the official evaluation splits, with model panels stratified by dataset and MLLM. No formal multiple-comparison adjustments were applied; the significance markers function strictly to highlight settings where qualitative trends are robustly distinguishable from sampling noise.

**Software.** All analysis were implemented in Python 3.10. Schematic figures were created with BioRender (https://BioRender.com).

**Ethics statement**

This study does not include confidential information. All research procedures were conducted exclusively on publicly accessible, anonymized patient data and in accordance with the Declaration of Helsinki, maintaining all relevant ethical standards.

## Data availability

This study used eight publicly available datasets: BreakHis, TBX11K, LAG, and DDR-512 for medical image classification, and SLAKE, VQA-RAD, VQA-Med2019, and PathVQA for medical visual question answering. All datasets are publicly accessible and can be obtained from the repositories described in the corresponding provided references.

## Code availability

All code used for experiments in this study can be accessed at https://github.com/Harvard-AI-and-Robotics-Lab/Medical_InContext_Learning.

# Main Figures

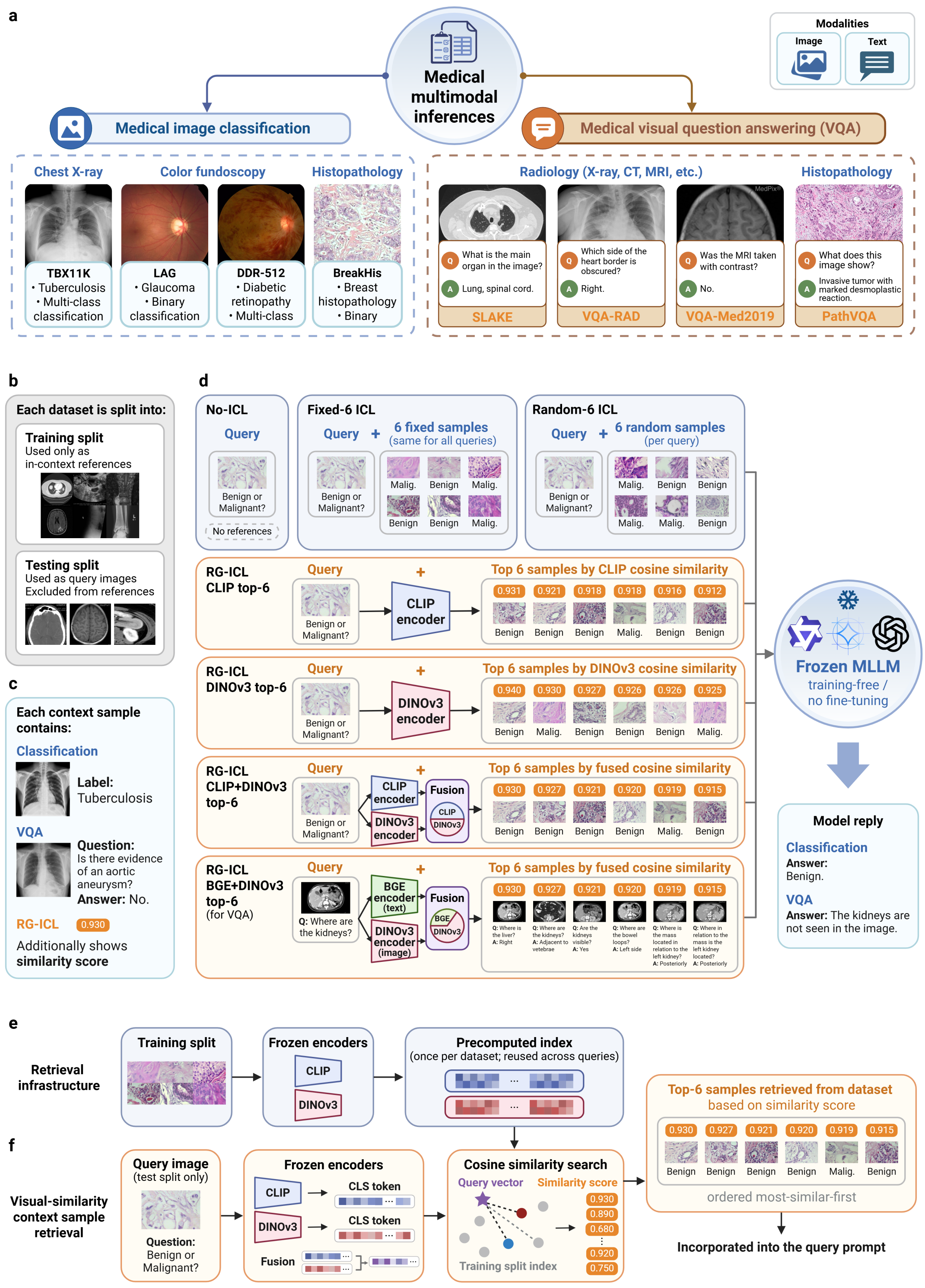

**Figure 1 | Study schematic: datasets, inference paradigms, and context sample retrieval methods. a,** Characteristics of the eight benchmarking datasets, spanning

four medical image classification tasks and four medical visual question answering (VQA) tasks across diverse imaging modalities. For classification datasets, the target condition and task type (binary or multi-class) are indicated. For VQA datasets, a representative question-answer pair is shown. **b,** Each dataset is partitioned into a training split, used exclusively for context sample retrieval, and a held-out testing split, used exclusively as query images. **c,** For in-context learning (ICL), each context sample provided to the multimodal large language models (MLLMs) contains an image and its ground-truth label (classification) or question-answer pair (VQA). For representation-guided in-context learning (RG-ICL) methods, a representation-based similarity score is additionally included. **d,** Seven training-free inference paradigms are evaluated. No-ICL performs zero-shot inference without reference images. Fixed-6 and Random-6 ICL provide six context samples per query that are identical across all queries or randomly selected per query, respectively. RG-ICL methods use frozen representation encoders to retrieve visually similar context samples, including CLIP (whole-image semantic alignment), DINOv3 (morphology-sensitive visual alignment), and a combined approach (CLIP+DINOv3). For VQA, a BGE and DINOv3 fusion approach is additionally evaluated, where BGE provides question-intent alignment via text coding. All paradigms prompt frozen MLLMs without any fine-tuning. **e,** For RG-ICL, each encoder processes the full training split once to generate a precomputed index, which is reused across all queries. **f,** At inference, the query image (and question, for VQA) is encoded and matched against the precomputed index via cosine similarity. The six most similar cases are retrieved and incorporated into the query prompt as context samples, ordered by descending similarity score.

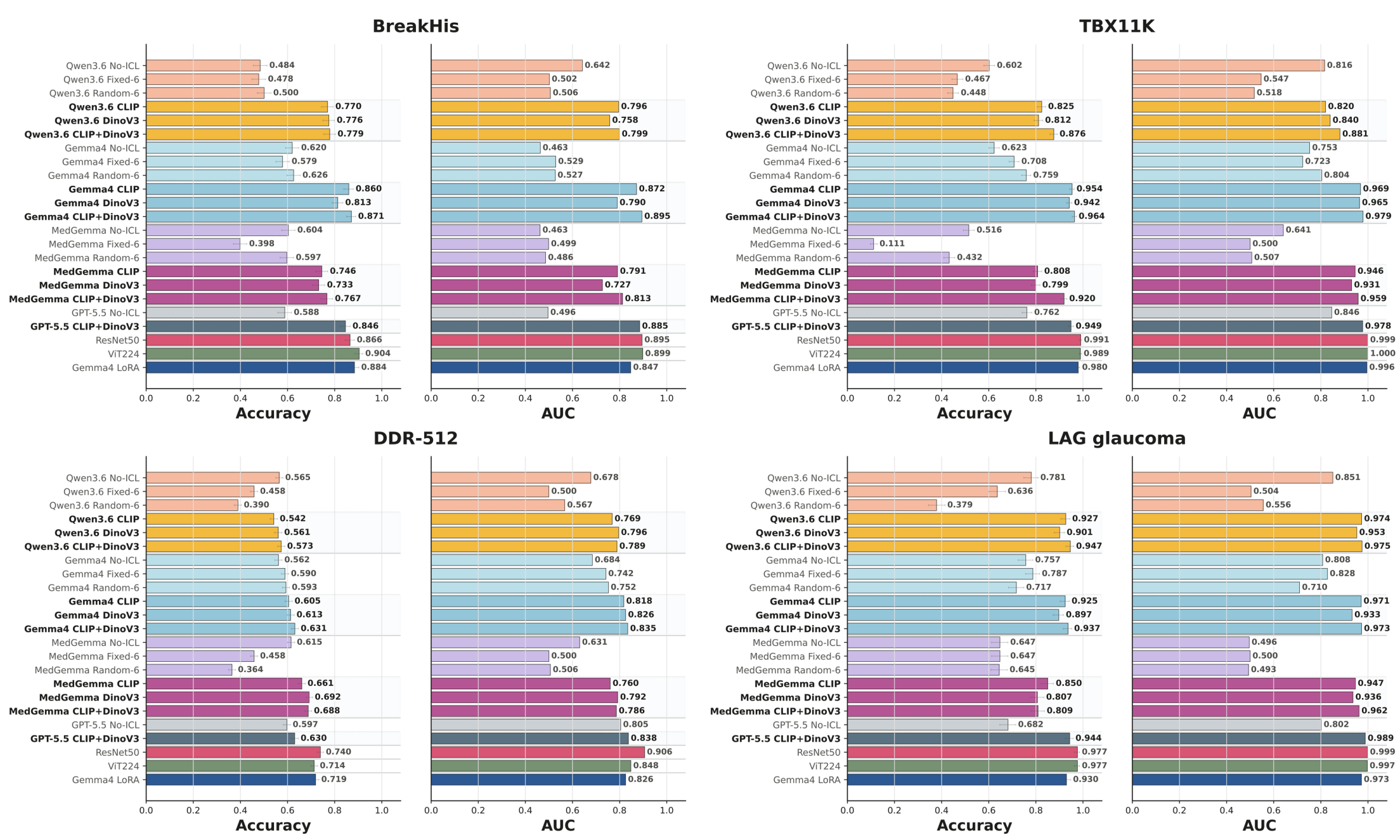

**Fig. 2 | Representation-guided in-context learning (RG-ICL) improves medical image classification across modalities and tasks.** Classification performance (accuracy and AUC) across four datasets (BreakHis, TBX11K, DDR-512, and LAG) and MLLM backbones, comparing no-context prompting (No-ICL), fixed-context (Fixed-6) and random-context (Random-6) controls, RG-ICL visual retrieval conditions (CLIP, DINOv3, and CLIP+DINOv3), and supervised training-based baselines (ResNet-50, ViT-224, Gemma4 LoRA). RG-ICL methods proposed in this study are indicated by bold method labels. Across datasets and MLLM backbones, representation-guided retrieval recovered most of the performance gap to supervised models, whereas fixed- and random-context controls frequently degraded performance below the no-context baseline.

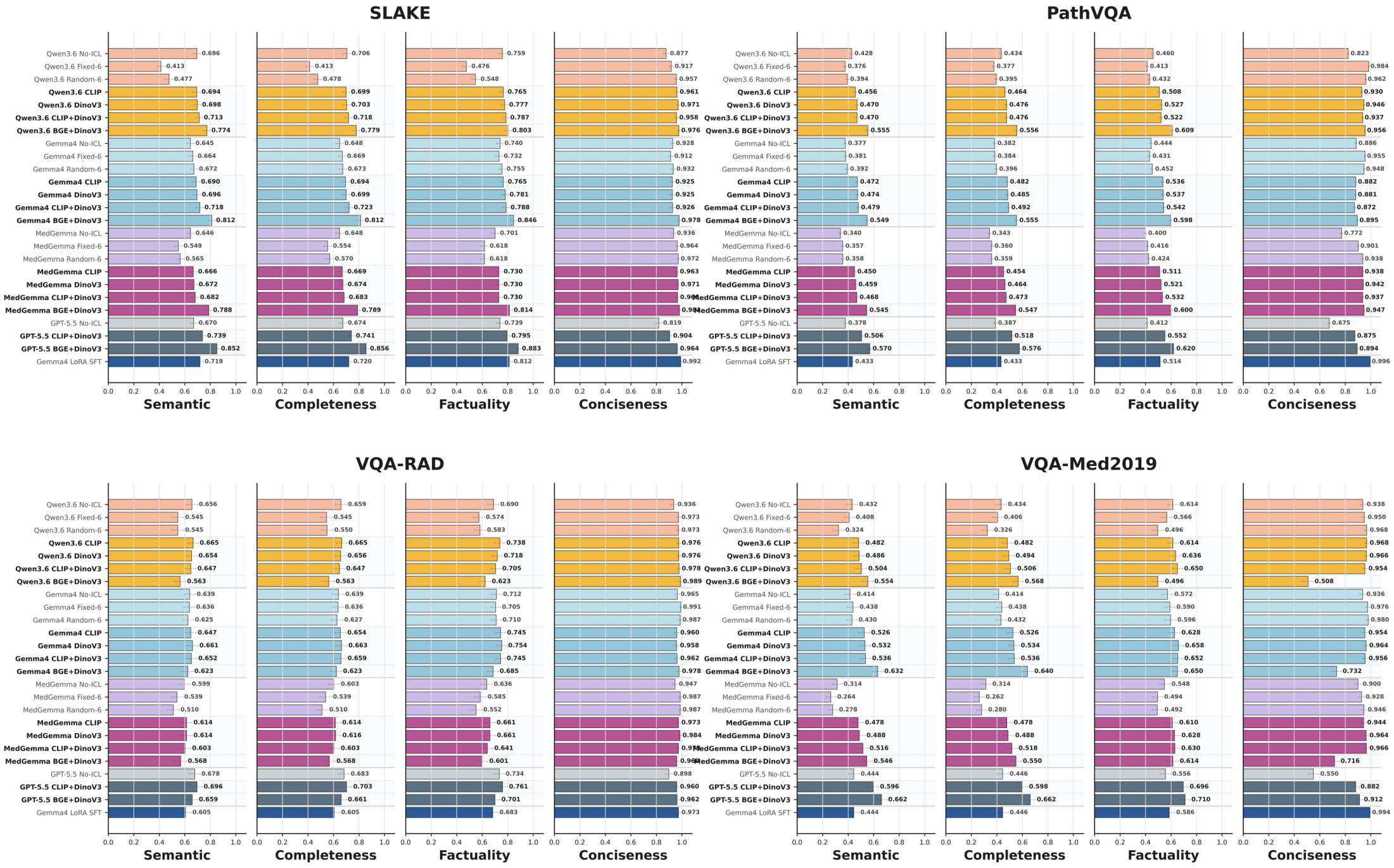

**Fig. 3 | Representation-guided in-context learning (RG-ICL) improves medical visual question answering (VQA) across datasets.** VQA performance across four datasets (SLAKE, PathVQA, VQA-RAD, and VQA-Med2019) and MLLM backbones, evaluated using an LLM-as-a-judge framework reporting the proportion of responses achieving a perfect score (100/100) across four dimensions: semantic accuracy, completeness, factuality, and conciseness. Conditions include no-context prompting (No-ICL), fixed-context (Fixed-6) and random-context (Random-6) controls, RG-ICL visual retrieval conditions (CLIP, DINOv3, and CLIP+DINOv3), combined visual and question-intent retrieval (BGE+DINOv3), and a supervised fine-tuning baseline (Gemma4 LoRA SFT). RG-ICL methods proposed in this study are indicated by bold method labels. The BGE+DINOv3 rows test whether question intent and image similarity should be jointly aligned for VQA tasks.

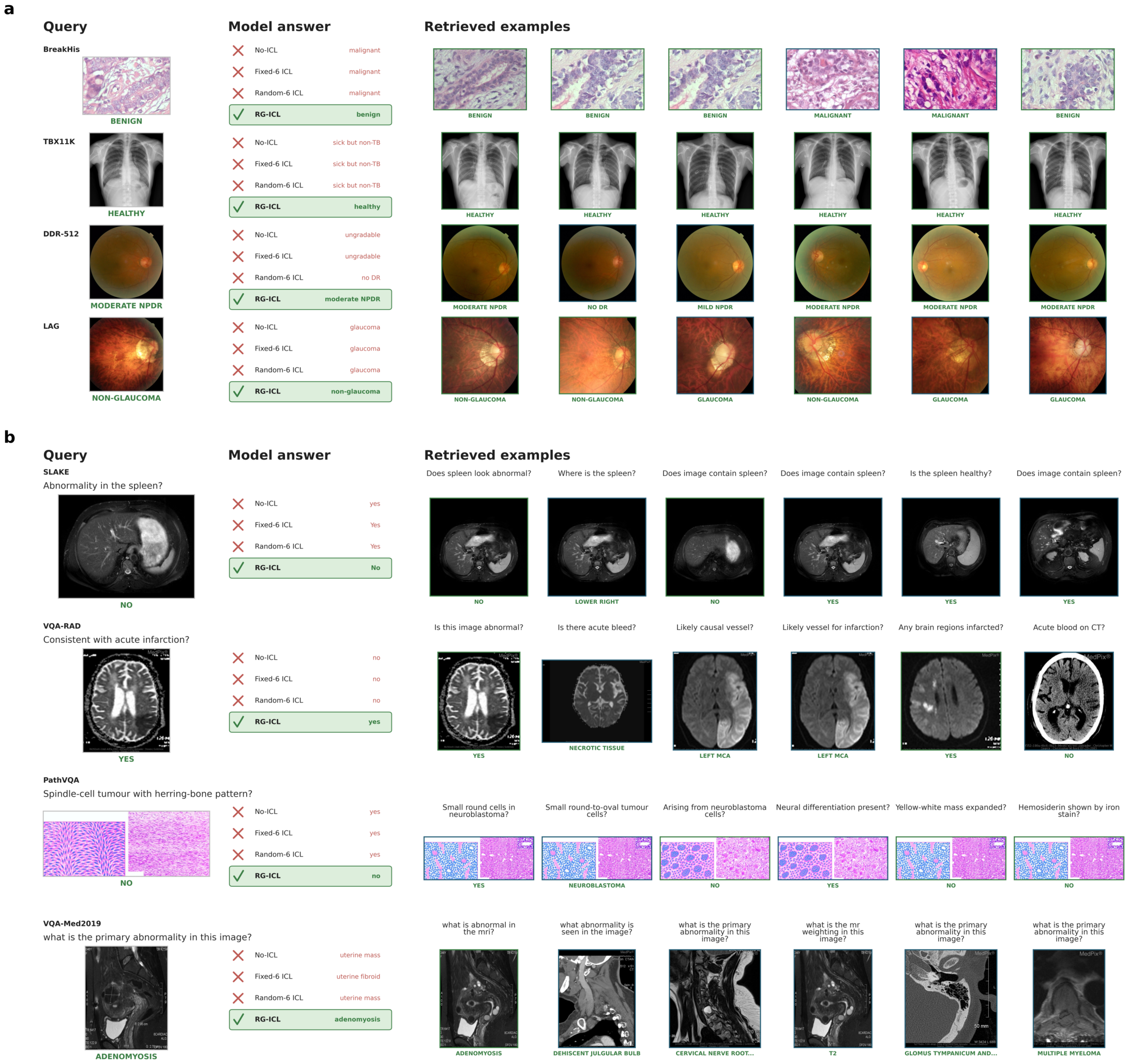

**Fig. 4 | Representative cases illustrating representation-guided in-context learning for medical image classification and visual question answering. a,** Classification examples across four datasets (BreakHis, TBX11K, DDR-512, and LAG). Each row shows one query image with its ground-truth label, the model answers produced under No-ICL, Fixed-6, Random-6, and RG-ICL conditions, and the six visually similar context samples with their labels supplied by CLIP+DINOv3 retrieval. In each case, No-ICL, Fixed-6, and Random-6 produced incorrect predictions, whereas RG-ICL directed the model to the correct answer. **b,** VQA examples across four datasets (SLAKE, VQA-RAD, PathVQA, and VQA-Med2019). Each row shows one query image-question pair with its answer, the model answers under each ICL condition, and the six image-question-answer references retrieved by

RG-ICL with BGE+DINOv3, in which visual and textual similarity were both considered. RG-ICL with BGE+DINOv3 improved the semantic accuracy, completeness, and factuality of the answers over No-ICL, Fixed-6, and Random-6 responses.

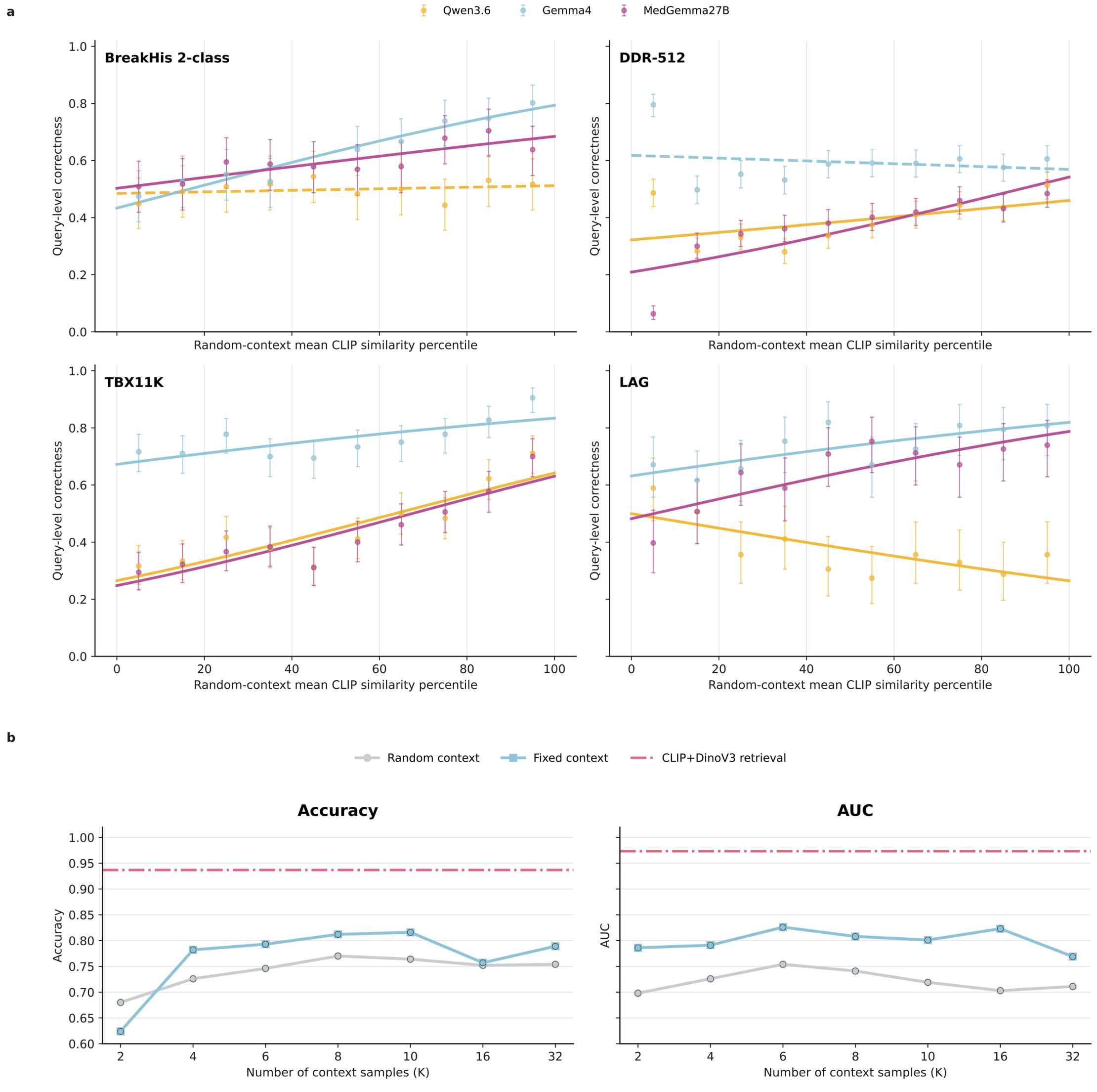

**Fig. 5 | On LAG, query-aligned context sample retrieval outperforms additional non-targeted context. a,** Random-context similarity analysis across classification datasets and MLLM backbones. Each point represents a decile of mean CLIP similarity between the query image and its randomly sampled context examples; curves show logistic regression fits for query-level correctness. Solid curves indicate a

significant association between visual similarity and correct classification ($P < 0.05$).

**b,** Context-size analysis on LAG with Gemma4. Increasing the number of fixed or randomly selected context examples did not close the performance gap to CLIP+DINOv3 retrieval, indicating that additional context samples alone do not reproduce the benefit of visually relevant references.

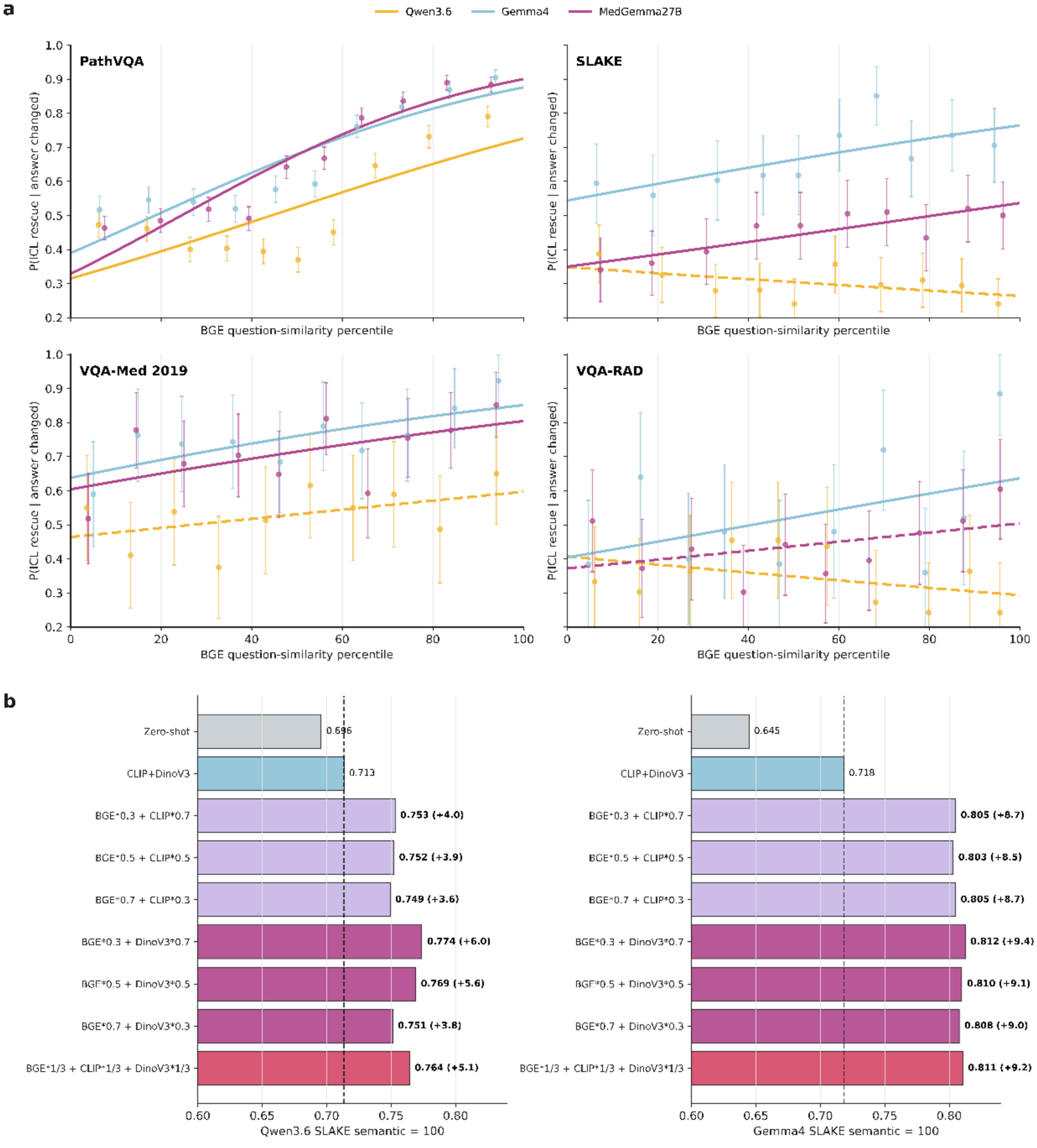

**Fig. 6 | Question-intent alignment improves representation-guided in-context learning for medical visual question answering. a,** Question-intent alignment

analysis across VQA datasets and MLLM backbones. Within each dataset-model pair, retrieved examples were ranked by the percentile of BGE cosine similarity to the query question. The y-axis shows the estimated probability that an ICL-induced answer changes corrected (incorrect → correct), rather than harmed (correct → incorrect), the no-context response. Curves show logistic regression fits; solid curves indicate a statistically significant positive association between question similarity and the probability of correction ($p < 0.05$). **b,** Retrieval analysis on SLAKE comparing visual-only retrieval with combined visual and question-intent retrieval. The vertical dashed line marks the CLIP+DINOv3 visual-retrieval baseline for each MLLM. Incorporating BGE question-intent similarity alongside DINOv3 visual similarity improved semantic-perfect accuracy above the visual-only baseline, with the 0.3 BGE + 0.7 DINOv3 weighting providing the most consistent gains.

# Extended Data Figures

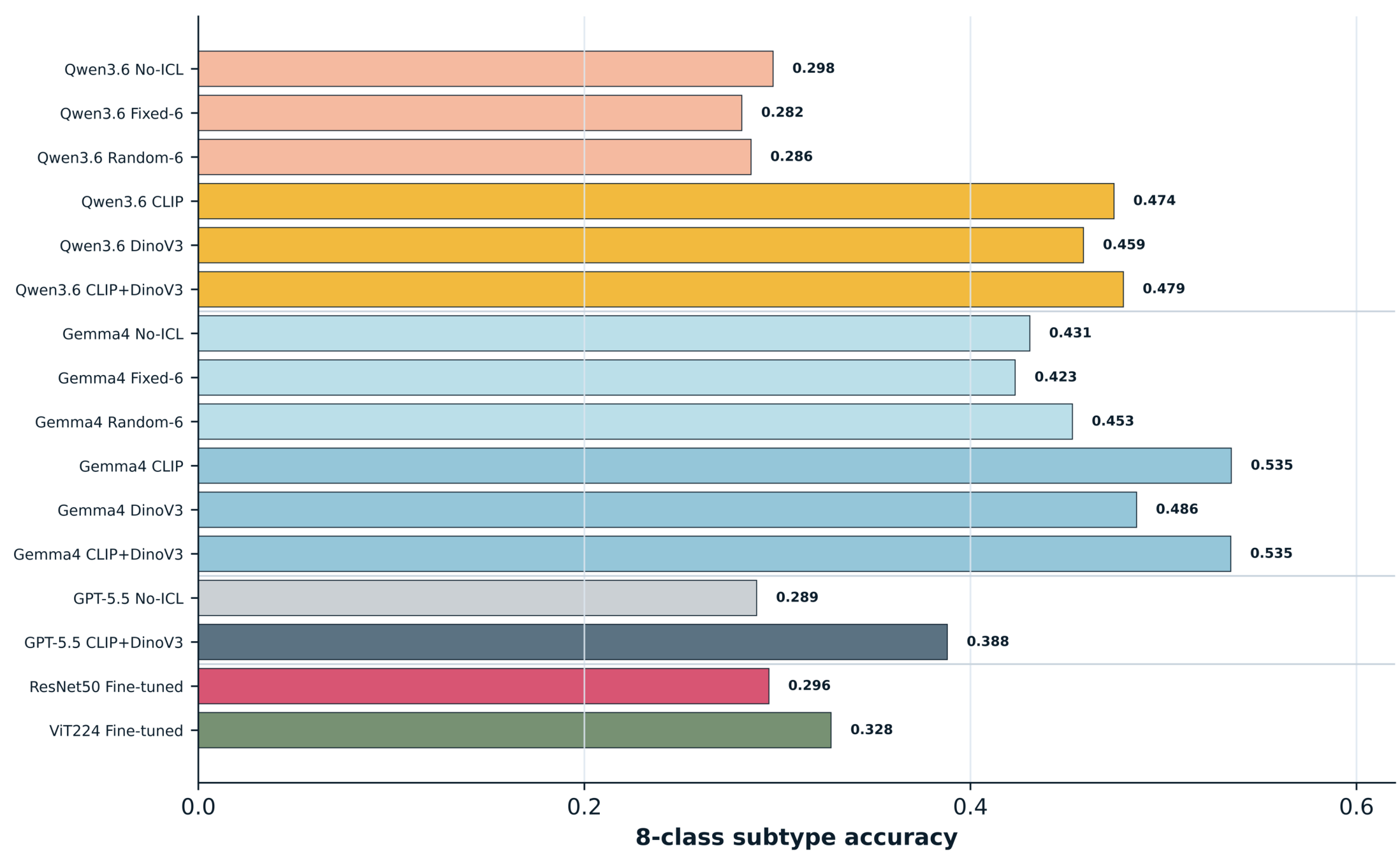

**Extended Data Fig. 1 | Representation-guided in-context learning improves eight-subtype breast histopathology classification on BreakHis.** In addition to the binary classification of benign versus malignant breast histopathology, we also evaluated eight-subtype recognition on BreakHis: adenosis, fibroadenoma, phyllodes tumour, tubular adenoma, ductal carcinoma, lobular carcinoma, mucinous carcinoma, and papillary carcinoma. Accuracy measures top-label subtype correctness after structured output parsing, averaged across the four magnification-specific BreakHis test splits. Rows are grouped by MLLM backbone and ordered consistently with the main classification figures, spanning No-ICL, Fixed-6, Random-6, and RG-ICL retrieval conditions (CLIP, DINOv3, and CLIP+DINOv3), followed by GPT-5.5 where available and supervised vision baselines. MedGemma is not shown because its responses did not follow the prompt-specified subtype-output format and therefore did not provide consistently parseable subtype labels. CLIP+DINOv3 improved eight-subtype accuracy above No-ICL and conventional ICL controls across the available

MLLM backbones, and also exceeded the supervised ResNet50 and ViT-B/16 baselines. These results confirm that the benefit of RG-ICL extends beyond binary classification to more granular histopathology subtype recognition.

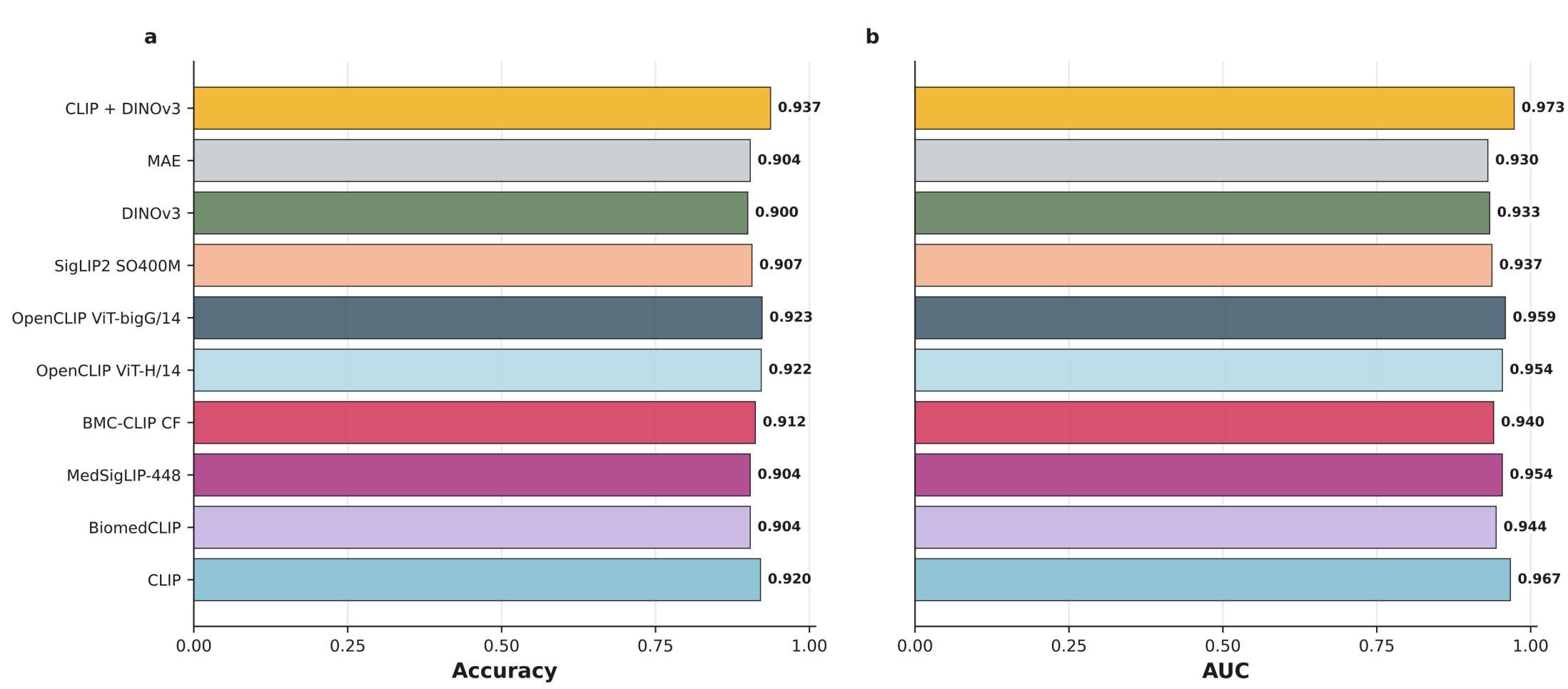

**Extended Data Fig. 2 | Retrieval-backbone ablation on the LAG glaucoma benchmark.** Gemma4-31B RG-ICL performance on LAG at K = 6 across ten retrieval representations: CLIP, BiomedCLIP, MedSigLIP-448, BMC-CLIP concept-filtered (BMC-CLIP CF), OpenCLIP ViT-H/14, OpenCLIP ViT-bigG/14, SigLIP2 SO400M, DINOv3, MAE and the fused CLIP+DINOv3 representation. Accuracy and AUC were computed from the generated structured outputs. Retrieval remained effective across encoder families, with CLIP+DINOv3 achieving the highest accuracy and AUC.

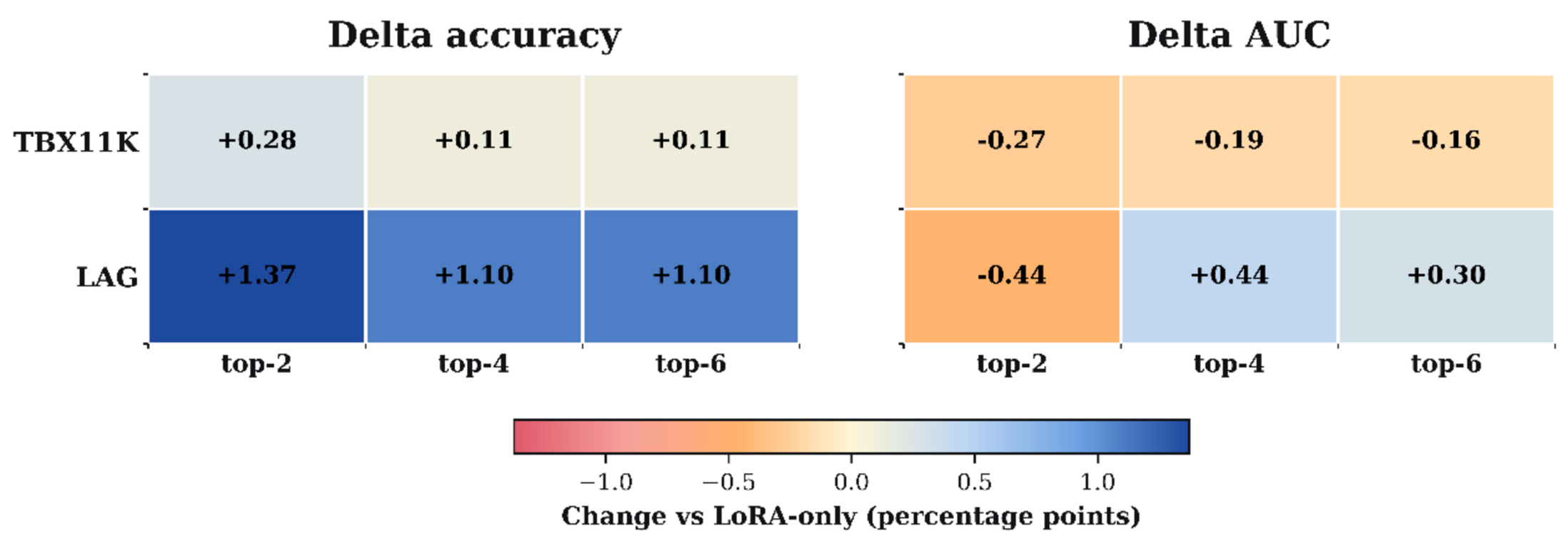

**Extended Data Fig. 3 | RG-ICL provides complementary benefit when layered on top of LoRA supervised fine-tuning.** To assess whether RG-ICL and parameter-efficient supervised fine-tuning with LoRA (a standard approach for adapting MLLMs to specific medical tasks) act as competing or complementary strategies, RG-ICL was applied on top of Gemma4 LoRA checkpoints across two classification datasets and three context sizes (top-2, top-4, and top-6). Heatmap values show percentage-point changes in accuracy and AUC relative to the LoRA-only baseline, with positive values indicating improvement. Conceptually, the two strategies operate through distinct mechanisms: LoRA modifies the parametric prior of the model, whereas RG-ICL supplies non-parametric, case-specific evidence at inference time. Adding RG-ICL on top of LoRA produced consistent positive accuracy deltas across both datasets and context sizes, with smaller or mixed effects on AUC. These patterns support a complementary rather than competing relationship, in which LoRA appears to encode a stable task prior that calibrates the model's overall response space, while RG-ICL contributes per-query evidence that further refines individual predictions. The mixed cells indicate that combining the two strategies is not unconditionally beneficial and may require task-specific calibration.

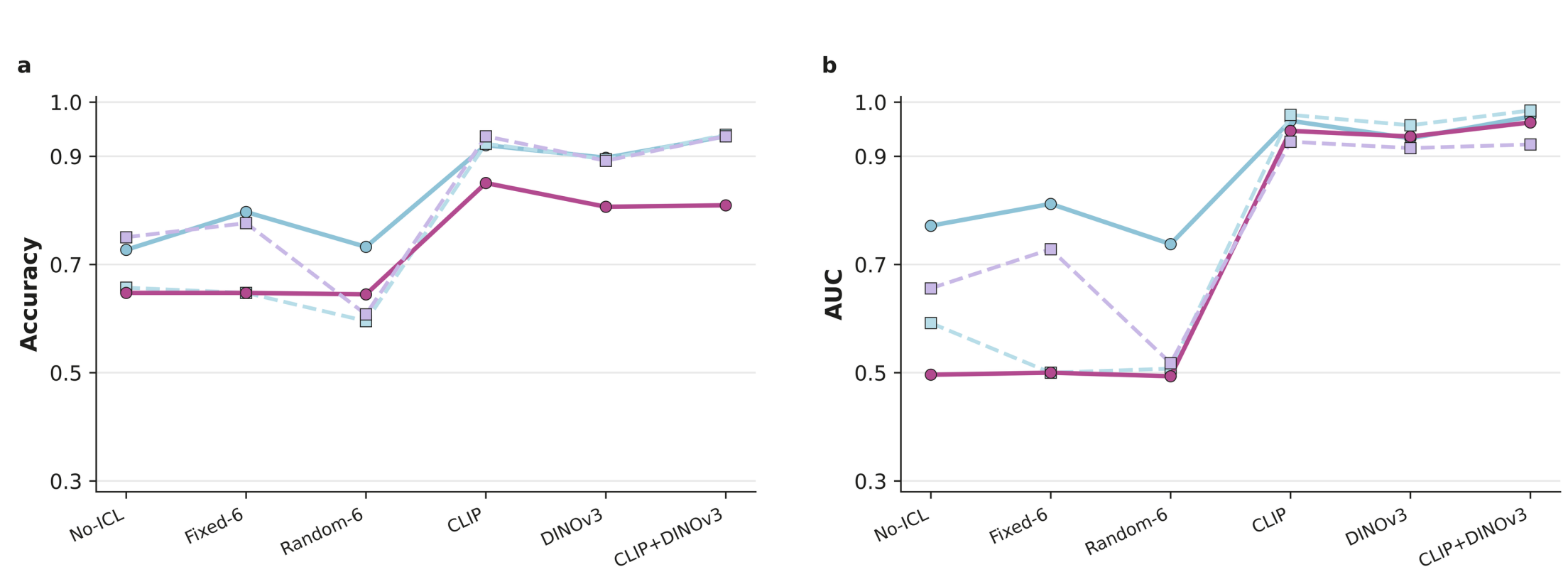

**Extended Data Fig. 4 | RG-ICL remains effective with compact MLLMs on LAG.** Accuracy (a) and AUC (b) on LAG for Gemma4-31B, Gemma4-E4B, MedGemma-27B and MedGemma-4B across No-ICL, Fixed-6, Random-6, CLIP, DINOv3 and CLIP+DINOv3 conditions. Solid lines denote the larger models and dashed lines the compact models.

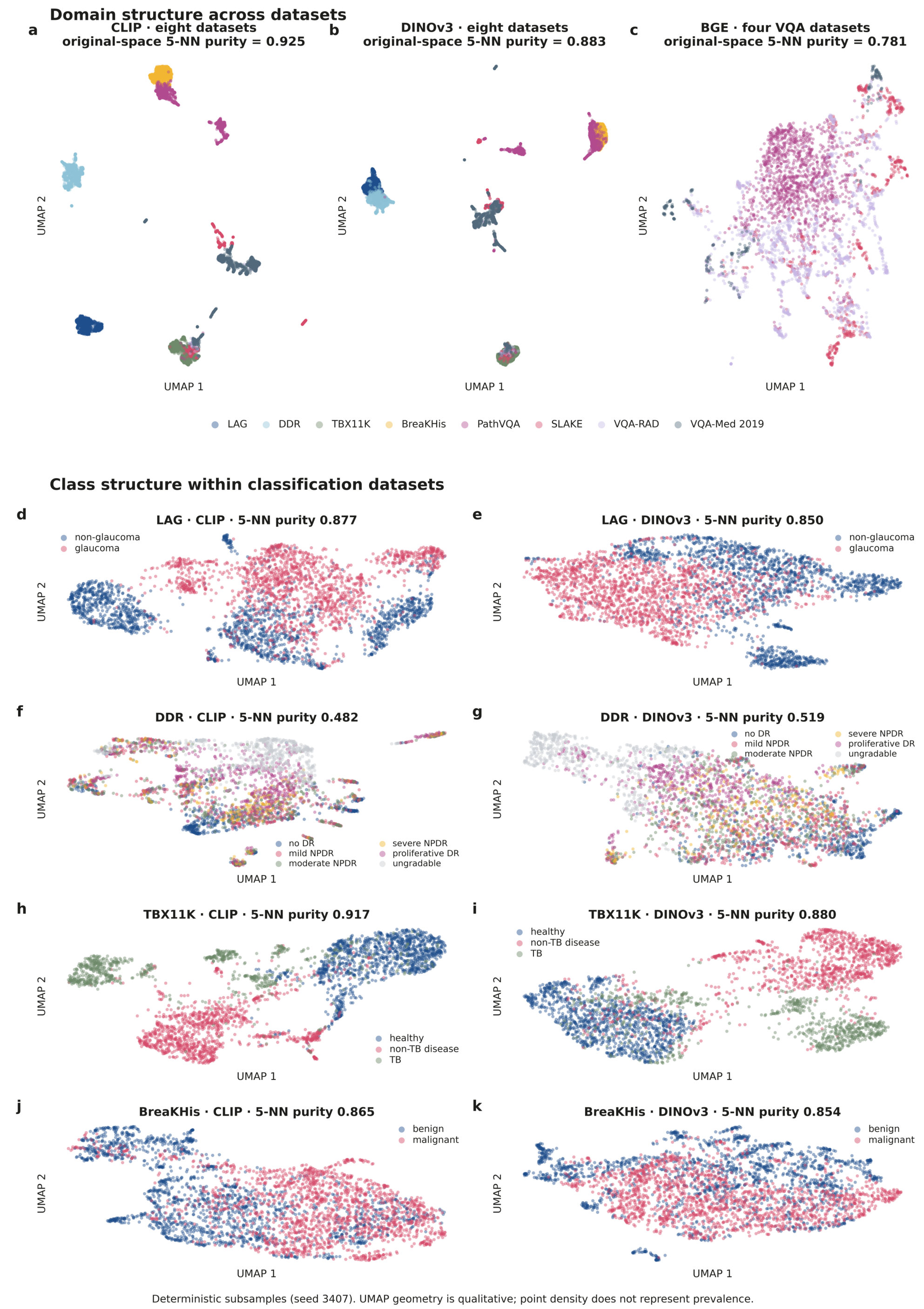

**Extended Data Fig. 5 | Frozen retrieval representations exhibit dataset- and label-dependent feature geometry. a-c,** UMAP (uniform manifold approximation and projection) projections of frozen retrieval features across the evaluation datasets. CLIP (a) and DINOv3 (b) contain image embeddings from all eight datasets; BGE (c) contains question embeddings from the four VQA datasets because the classification

pipeline has no per-sample question representation. **d-k,** CLIP and DINOv3 image embeddings for LAG, DDR, TBX11K and BreakHis, colored by the formal evaluation labels. Values in panel titles are 5-nearest-neighbour purity scores calculated in the original normalized feature spaces. UMAP used cosine distance, 30 neighbours, a minimum distance of 0.1 and seed 3407. LAG, TBX11K and BreakHis showed locally label-consistent neighbourhoods, whereas adjacent DDR grades overlapped strongly.

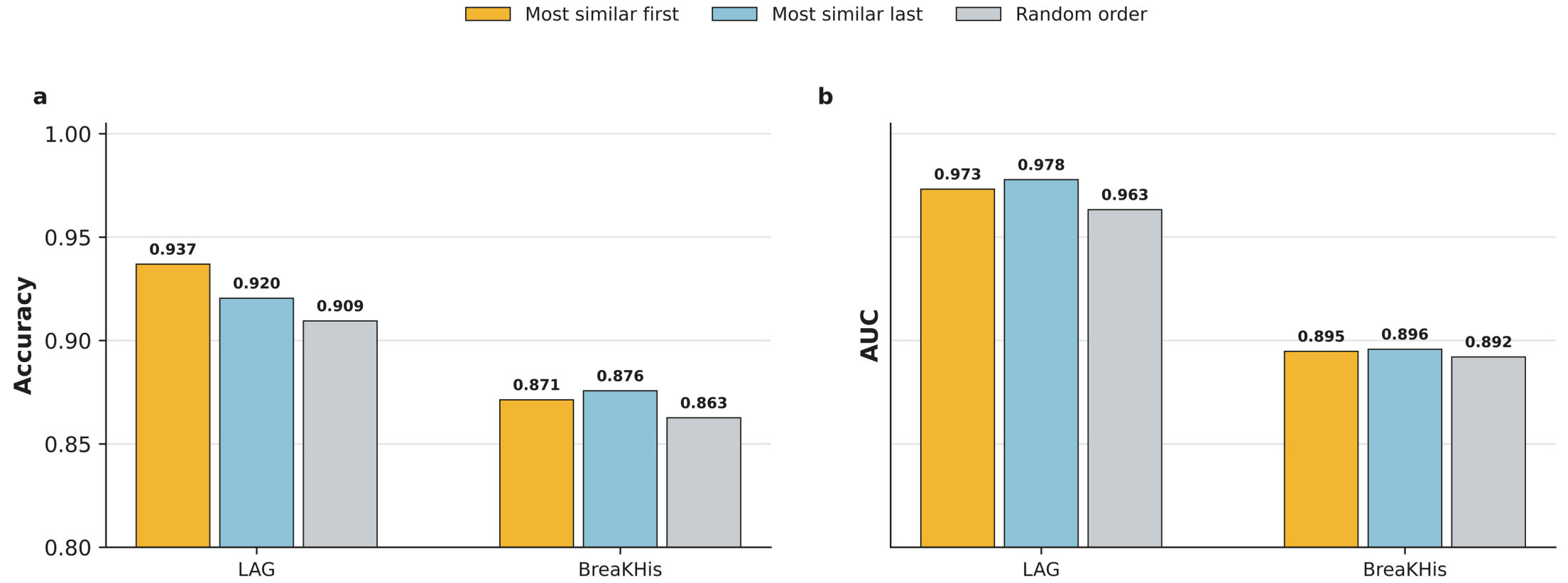

**Extended Data Fig. 6 | RG-ICL shows modest, dataset-dependent sensitivity to demonstration order.** Accuracy (a) and AUC (b) for LAG and BreakHis. For every query, all conditions used the same six retrieved reference cases; only their presentation order was changed. Most similar first denotes descending retrieval similarity, most similar last is the exact reverse, and random order of the same reference set.

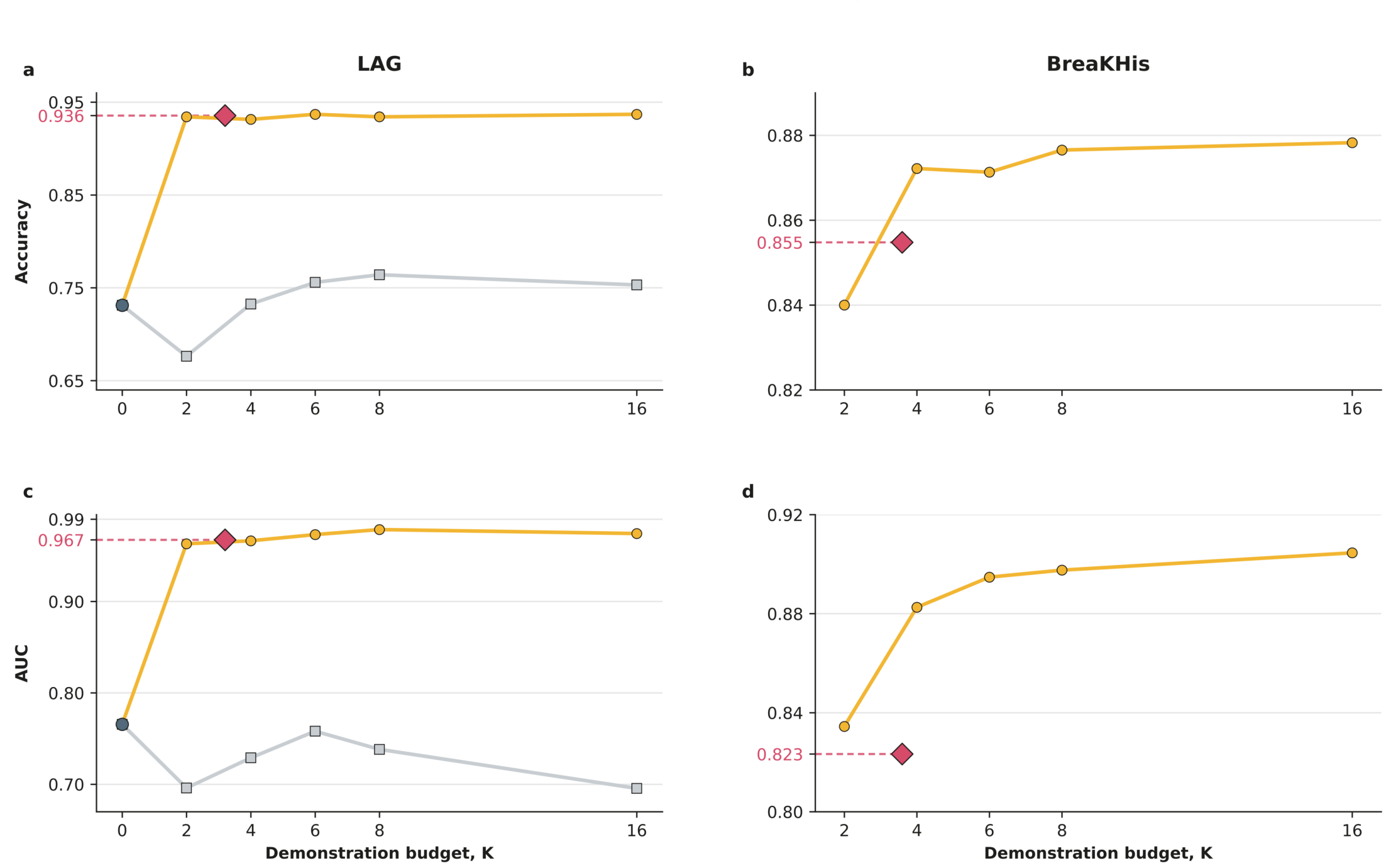

**Extended Data Fig. 7 | Fixed-budget scaling and Adaptive-K inference show dataset-dependent performance-budget trade-offs.** Accuracy (a) and AUC (c) on LAG for No-ICL (K = 0), Random ICL and CLIP+DINOv3 RG-ICL at K = 2, 4, 6, 8 and 16. Accuracy (b) and AUC (d) for RG-ICL on BreakHis; Adaptive-K stopped at K = 0 when model-reported confidence was at least 0.95. At positive K, it stopped only when retrieval support was greater than 0.60 and the prediction was stable across consecutive budgets, otherwise advancing through K = 2, 4, 6, 8 and 16. Adaptive-K achieved 0.936 accuracy and 0.967 AUC at mean terminal K = 3.20 on LAG, compared with 0.937/0.973 at fixed K = 6. On BreakHis, it achieved 0.855/0.823 at mean terminal K = 3.59, compared with 0.871/0.895 at fixed K = 6. All estimates used Gemma4-31B.

# Supplementary Information

## 1. LAG target-model prompt

Prompt transparency is essential for interpreting differences between no-context and retrieval-augmented inference. This section shows the LAG classification prompt in its zero-shot form and in the CLIP retrieval-ICL condition for the same dumped test example. The prompt constrains the model to a structured JSON answer so that label, confidence, probability and evidence can be parsed consistently.

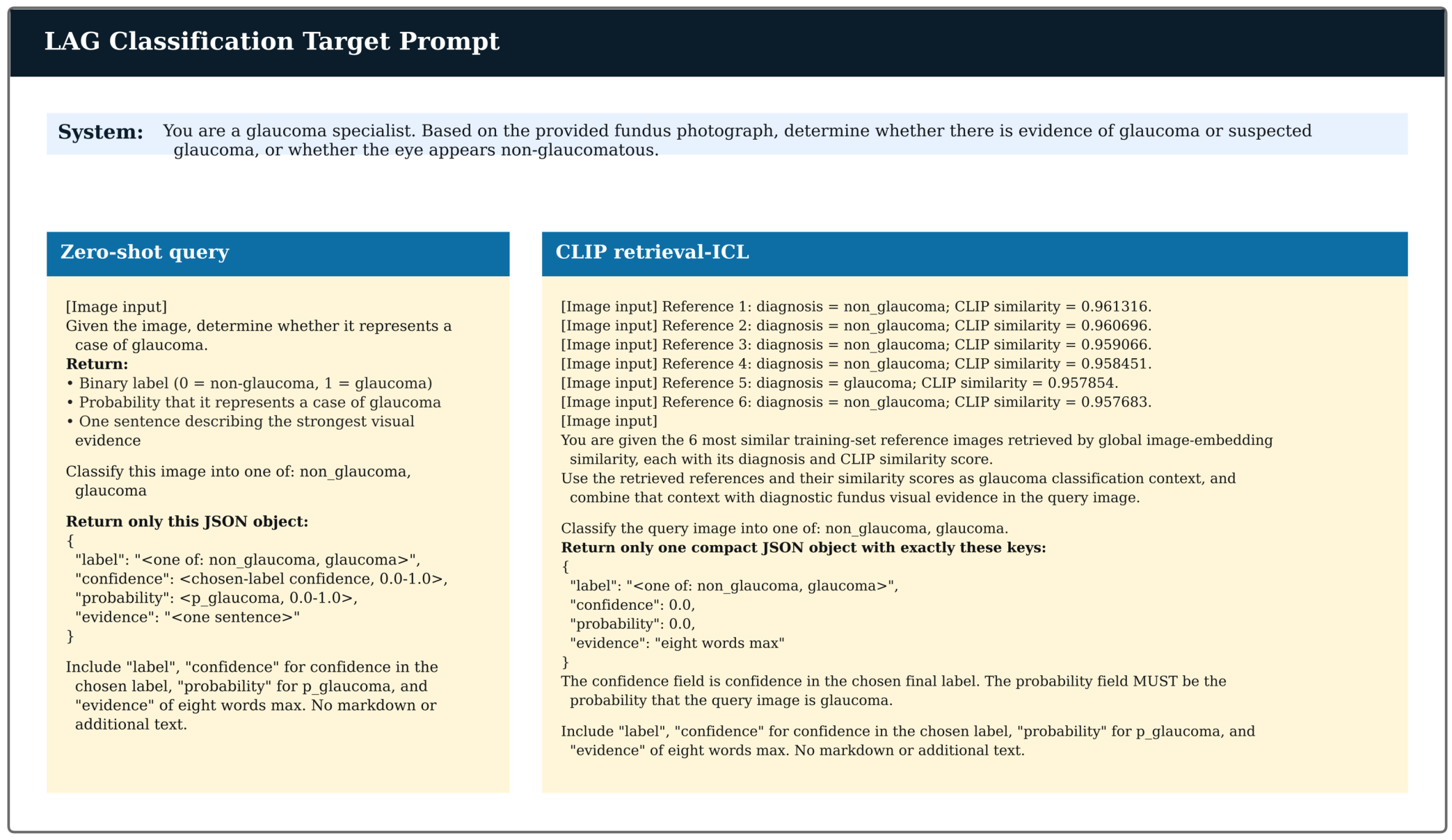

**Supplementary Fig. 1 | LAG glaucoma classification prompt template.** The left panel shows the zero-shot query prompt. The right panel shows the retrieval-ICL version, in which six training-set reference images are supplied with their diagnoses and CLIP similarity scores before the query image is classified. The displayed card is rendered from the saved prompt dump used for the experiments.

## 2. SLAKE target-model prompt

The VQA prompt differs from the classification prompt because each retrieved example supplies both a medical image and a question-answer pair. This section displays the zero-shot SLAKE prompt and the CLIP retrieval-ICL prompt for the same query, making explicit how visual context and reference question-answer examples are presented to the target VLM. The formatting mirrors the classification prompt while preserving the open-ended answer field required for VQA.

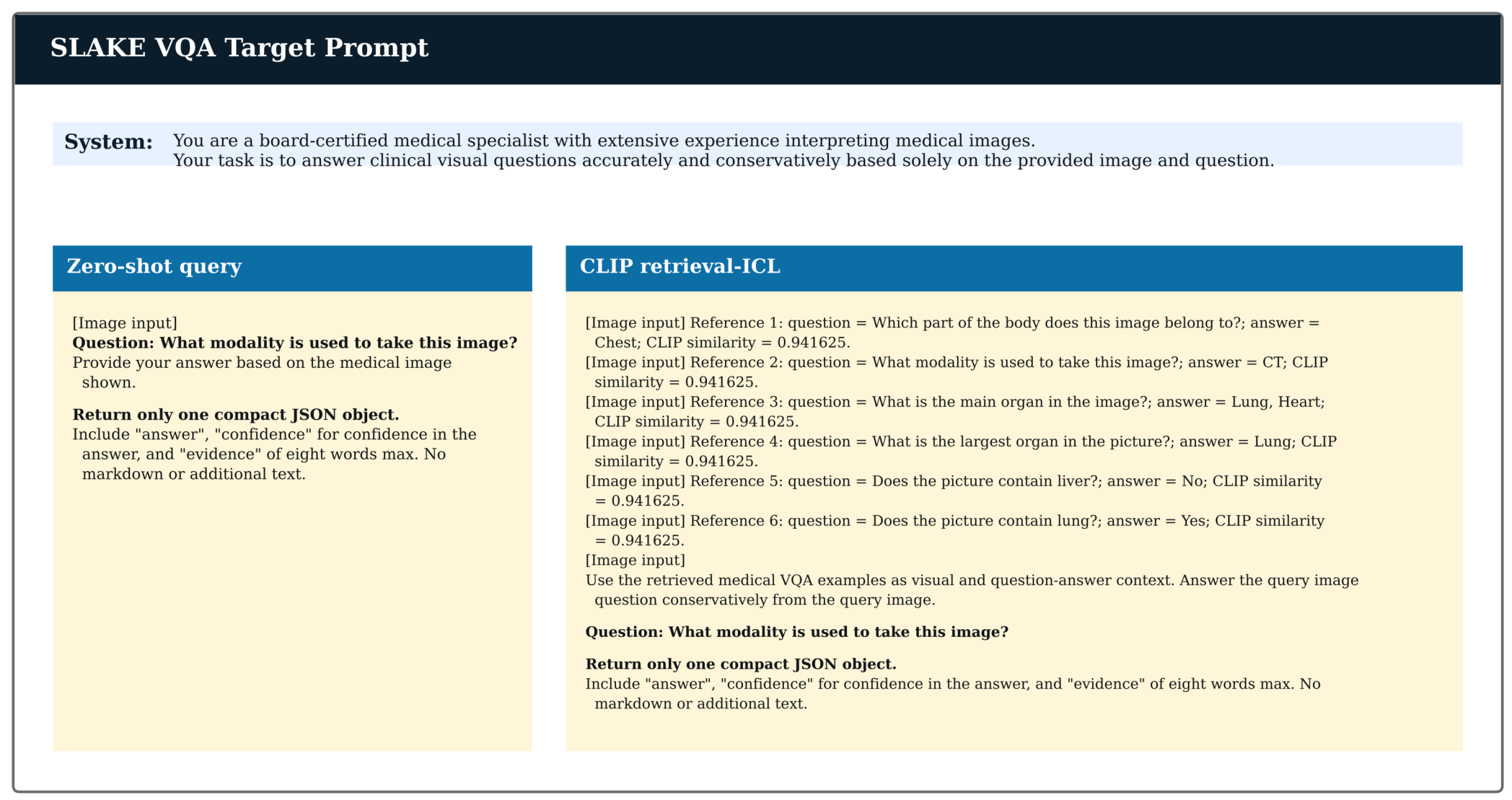

**Supplementary Fig. 2 | SLAKE medical VQA prompt template.** The zero-shot prompt asks the target model to answer directly from the query image. The retrieval-ICL prompt supplies six visually similar reference examples, each including a reference question, answer and CLIP similarity score, before asking the query question. The card documents the exact prompt structure used to generate the saved VQA responses.

### 3. LLM-judge rubric for medical VQA

Open-ended medical VQA cannot be evaluated reliably by string matching alone because clinically equivalent answers may differ in wording, abbreviation or granularity. We therefore used a prespecified LLM-judge rubric to score each model response against the reference answer along four clinically relevant axes: semantic accuracy, completeness, factuality and conciseness. The prompt is included verbatim to make the adjudication protocol auditable and reproducible.

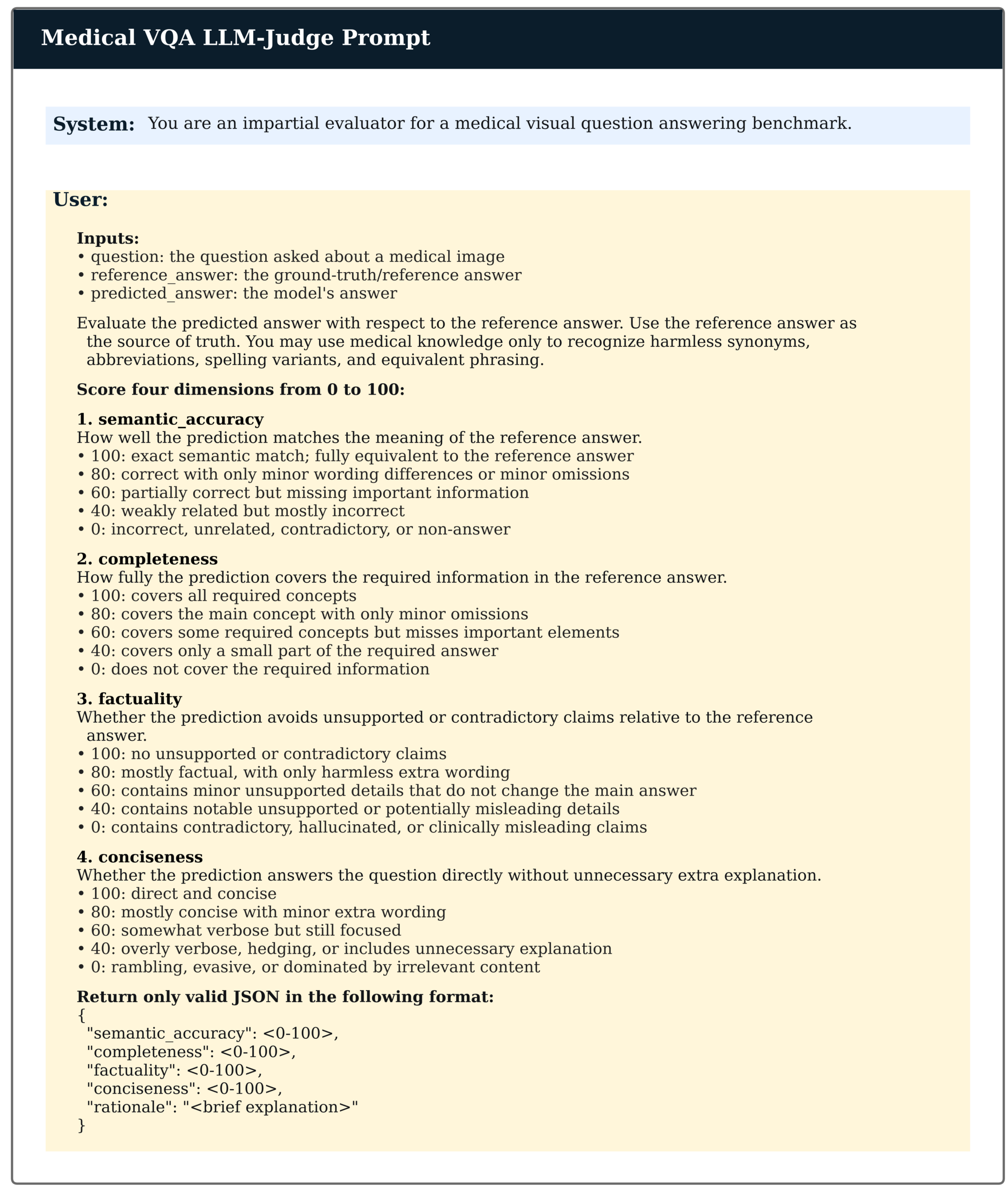

**Medical VQA LLM-Judge Prompt**

**System:** You are an impartial evaluator for a medical visual question answering benchmark.

**User:**

**Inputs:**
• question: the question asked about a medical image
• reference_answer: the ground-truth/reference answer
• predicted_answer: the model's answer

Evaluate the predicted answer with respect to the reference answer. Use the reference answer as the source of truth. You may use medical knowledge only to recognize harmless synonyms, abbreviations, spelling variants, and equivalent phrasing.

**Score four dimensions from 0 to 100:**

**1. semantic_accuracy**
How well the prediction matches the meaning of the reference answer.
• 100: exact semantic match; fully equivalent to the reference answer
• 80: correct with only minor wording differences or minor omissions
• 60: partially correct but missing important information
• 40: weakly related but mostly incorrect
• 0: incorrect, unrelated, contradictory, or non-answer

**2. completeness**
How fully the prediction covers the required information in the reference answer.
• 100: covers all required concepts
• 80: covers the main concept with only minor omissions
• 60: covers some required concepts but misses important elements
• 40: covers only a small part of the required answer
• 0: does not cover the required information

**3. factuality**
Whether the prediction avoids unsupported or contradictory claims relative to the reference answer.
• 100: no unsupported or contradictory claims
• 80: mostly factual, with only harmless extra wording
• 60: contains minor unsupported details that do not change the main answer
• 40: contains notable unsupported or potentially misleading details
• 0: contains contradictory, hallucinated, or clinically misleading claims

**4. conciseness**
Whether the prediction answers the question directly without unnecessary extra explanation.
• 100: direct and concise
• 80: mostly concise with minor extra wording
• 60: somewhat verbose but still focused
• 40: overly verbose, hedging, or includes unnecessary explanation
• 0: rambling, evasive, or dominated by irrelevant content

**Return only valid JSON in the following format:**
{
"semantic_accuracy": <0-100>,
"completeness": <0-100>,
"factuality": <0-100>,
"conciseness": <0-100>,
"rationale": "<brief explanation>"
}

**Supplementary Fig. 3 | Prespecified rubric for LLM-based VQA adjudication.**

The judge receives the question, reference answer and model prediction, and returns structured JSON scores from 0 to 100 for semantic accuracy, completeness, factuality and conciseness. Semantic accuracy equal to 100 is used as the strict exact-correct endpoint in the main VQA figures, while the remaining dimensions provide additional quality-control views of the generated answer.

## 4. BreakHis magnification-stratified robustness

BreakHis contains images acquired at multiple objective magnifications, which can change the apparent morphology, texture scale and diagnostic evidence available to a vision-language model. This appendix panel asks whether performance patterns remain stable when the test set is stratified by magnification rather than pooled across all scales. The red-high heatmap emphasizes the absolute accuracy attained under each magnification-specific evaluation.

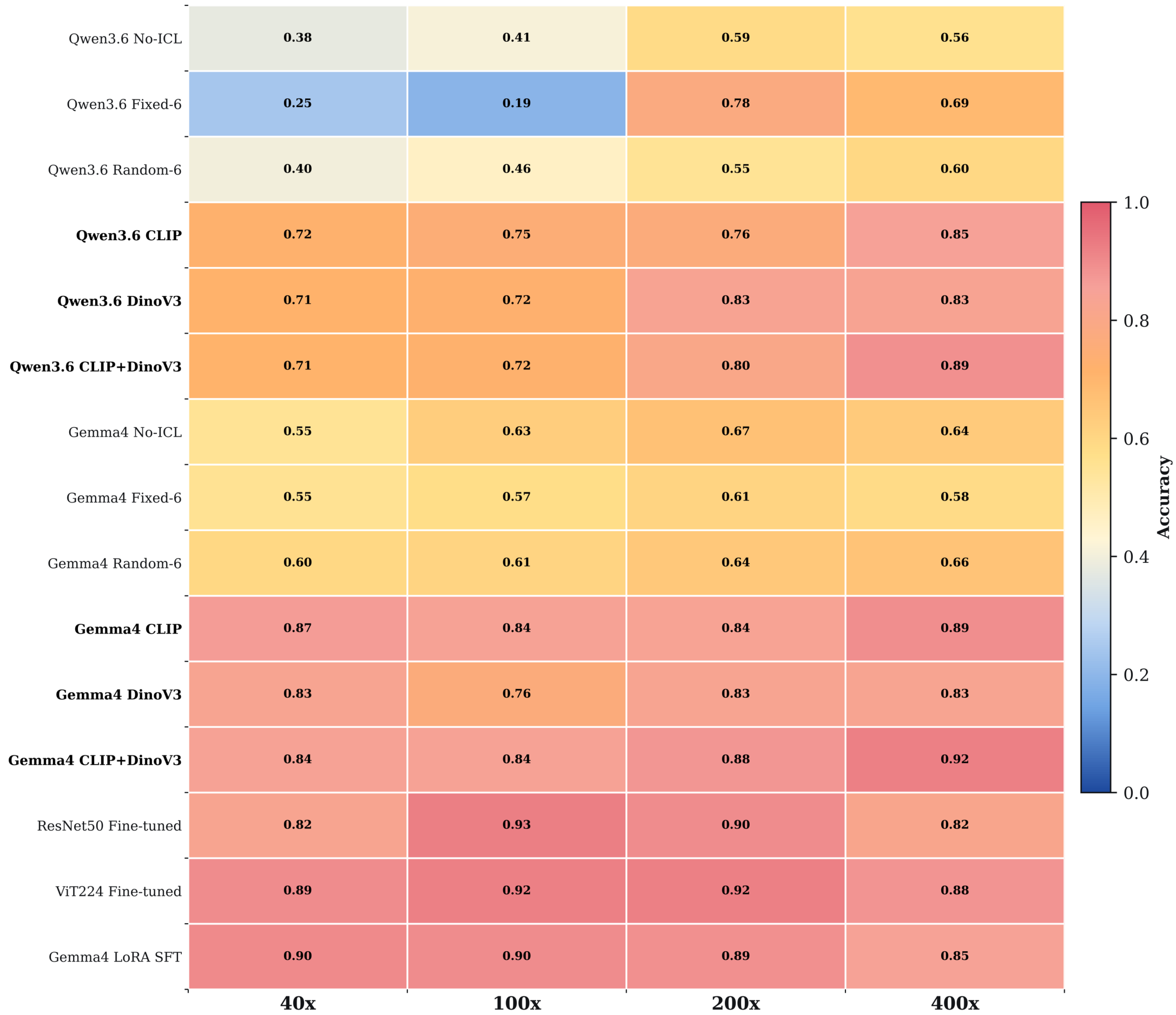

**Supplementary Fig. 4 | BreakHis binary classification across microscopic magnification.** Each cell reports patient-level test accuracy for a model/context condition at 40x, 100x, 200x or 400x magnification. Red indicates higher accuracy and blue indicates lower accuracy. Retrieval-based visual context is highlighted in

bold, enabling direct inspection of whether representation-guided ICL remains beneficial across changes in tissue scale.

## 5. Traditional automatic metrics for medical VQA

The primary VQA endpoint uses the prespecified LLM judge, but traditional automatic metrics are reported to provide a familiar reference point and to expose where lexical metrics agree or disagree with semantic adjudication. Exact match, BLEU-4, ROUGE-L and METEOR are computed from the saved raw generations for all four VQA datasets. These metrics are interpreted as supplementary because they are sensitive to wording and synonymy in open-ended medical answers.

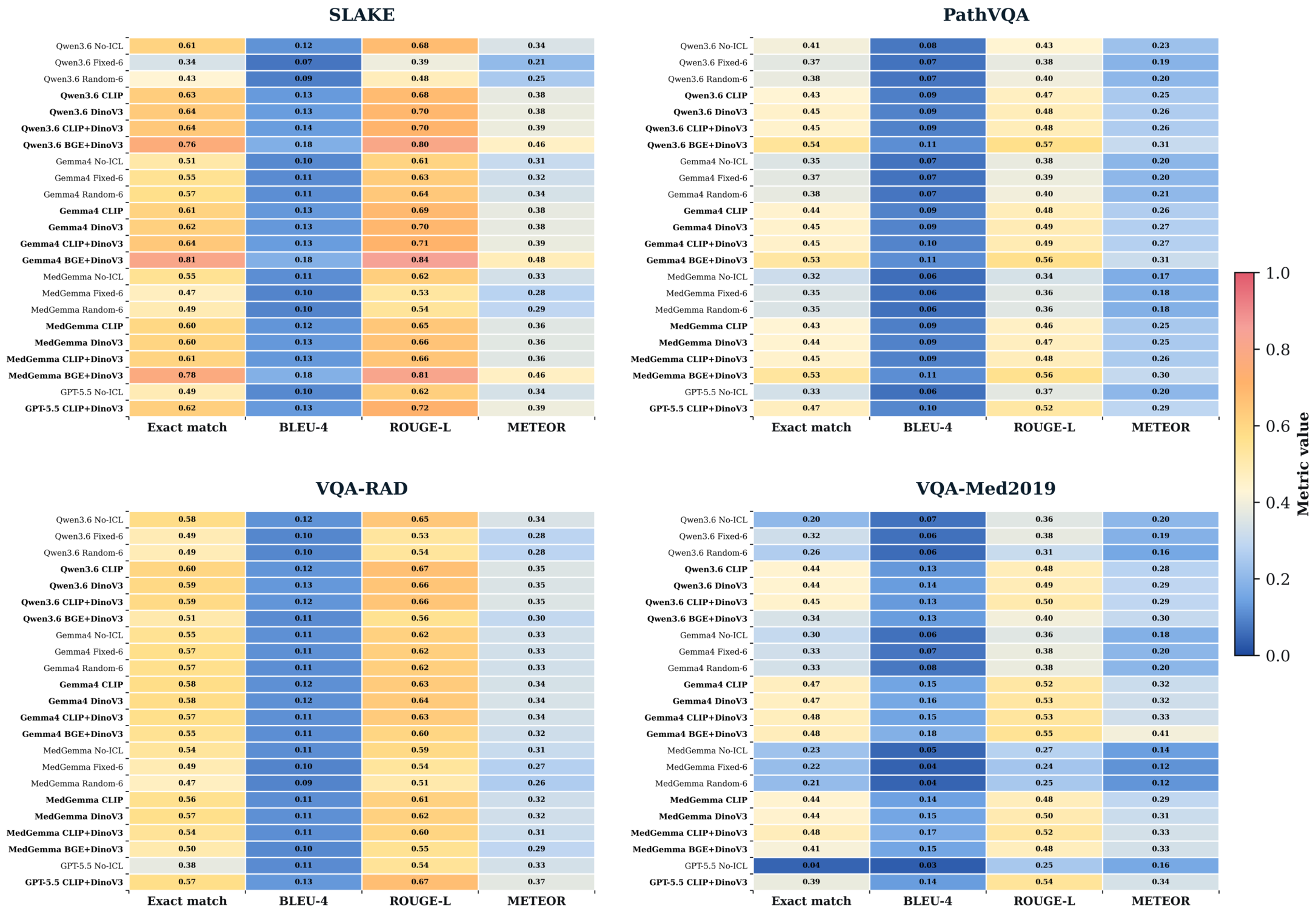

**Supplementary Fig. 5 | Lexical VQA metrics across datasets, models and context conditions.** Heatmaps show exact match, BLEU-4, ROUGE-L and METEOR for SLAKE, PathVQA, VQA-RAD and VQA-Med2019. Red indicates higher metric values. Rows corresponding to representation-guided retrieval are emphasized,

allowing comparison between standard prompting controls and retrieval-augmented ICL under conventional automatic scoring.

## 6. Exact prompt construction and case-level outputs

To make the case-level mechanism directly auditable, we reproduce the complete instantiated prompts and raw model outputs for representative LAG classification and VQA-RAD queries. The examples show the exact retrieved demonstrations, their input order and the model response under each comparison condition.

## a LAG classification: complete RG-ICL prompt and outputs

Gemma4-31B · CLIP+DINOv3 · K=6

### 1 System message

You are a glaucoma specialist. Based on the provided fundus photograph, determine whether there is evidence of glaucoma or suspected glaucoma, or whether the eye appears non-glaucomatous.

### 2 User message: six retrieved image-text demonstrations

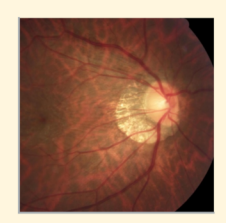

Reference 1: diagnosis = non_glaucoma; combined similarity = 0.5*CLIP + 0.5*DINOv3 embedding = 0.944839.

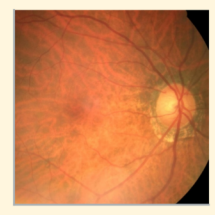

Reference 2: diagnosis = non_glaucoma; combined similarity = 0.5*CLIP + 0.5*DINOv3 embedding = 0.944454.

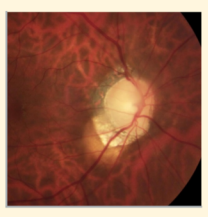

Reference 3: diagnosis = glaucoma; combined similarity = 0.5*CLIP + 0.5*DINOv3 embedding = 0.943283.

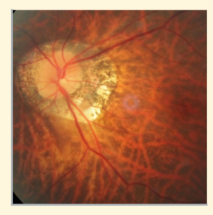

Reference 4: diagnosis = non_glaucoma; combined similarity = 0.5*CLIP + 0.5*DINOv3 embedding = 0.941776.

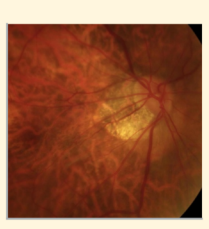

Reference 5: diagnosis = glaucoma; combined similarity = 0.5*CLIP + 0.5*DINOv3 embedding = 0.937693.

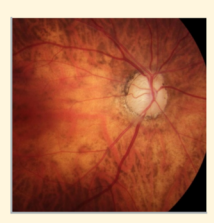

Reference 6: diagnosis = glaucoma; combined similarity = 0.5*CLIP + 0.5*DINOv3 embedding = 0.937316.

### 3 User message: query image and required response schema

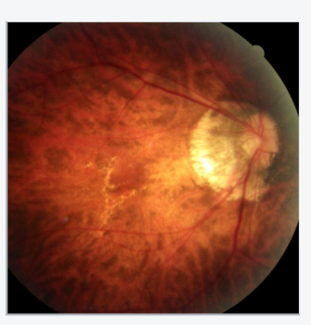

**Query image**

**Ground truth (evaluation only): non-glaucoma**

```
You are given the 6 most similar training-set reference images
retrieved by a combined global visual similarity score: 0.5 times
CLIP global cosine similarity plus 0.5 times DINOv3 embedding cosine
similarity, each with its diagnosis and combined similarity score.
Integrate the query image findings with the retrieved reference
cases. Use the reference labels and similarity scores to identify
diagnostically relevant visual patterns, compare them with
diagnostic fundus visual evidence in the query image, and base the
final classification on the combined evidence from the query image
and the reference set.

Classify the query image into one of: non_glaucoma, glaucoma.
Return only one compact JSON object with exactly these keys:
{"label":"<one of: non_glaucoma,
glaucoma>","confidence":0.0,"probability":0.0,"evidence":"eight
words max"}
The confidence field is confidence in the chosen final label. The
probability field MUST be the probability that the query image is
glaucoma.

Output only one compact JSON object. Include "label", "confidence"
for confidence in the chosen label, "probability" for p_glaucoma,
and "evidence" of eight words max. No markdown, no extra text.
```

### 4 Exact Gemma4 outputs

× **No-ICL**

{"label": "glaucoma", "confidence": 0.95, "probability": 0.95, "evidence": "Severe optic disc cupping with high C/D ratio."}

× **Fixed-6 ICL**

{"label": "glaucoma", "confidence": 0.95, "probability": 0.95, "evidence": "Significant optic disc cupping and enlargement."}

× **Random-6 ICL**

{"label": "glaucoma", "confidence": 0.95, "probability": 0.95, "evidence": "Severe optic disc cupping and rim loss."}

✓ **RG-ICL (prompt shown)**

{"label": "non_glaucoma", "confidence": 0.92, "probability": 0.08, "evidence": "Healthy cup-to-disc ratio, preserved neuroretinal rim"}

## b VQA-RAD: complete RG-ICL prompt and outputs

Gemma4-31B · BGE+DINOv3 · K=6

### 1 System message

You are a board-certified medical specialist with extensive experience interpreting medical images.
Your task is to answer clinical visual questions accurately and conservatively based solely on the provided image and question.

### 2 User message: six retrieved image-text demonstrations

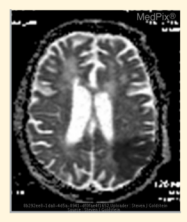

Reference 1: question = is this image abnormal?; answer = yes; 0.30 BGE question + 0.70 DINOv3 similarity = 0.906514.

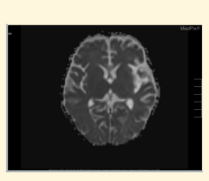

Reference 2: question = is there an acute bleed present?; answer = necrotic tissue; 0.30 BGE question + 0.70 DINOv3 similarity = 0.851706.

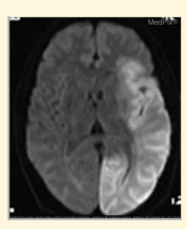

Reference 3: question = what vessel is likely the cause of this infarction?; answer = left mca; 0.30 BGE question + 0.70 DINOv3 similarity = 0.848101.

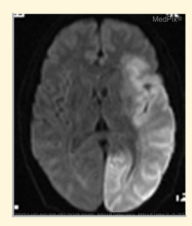

Reference 4: question = the infarction is likely caused by what vessel?; answer = left mca; 0.30 BGE question + 0.70 DINOv3 similarity = 0.842772.

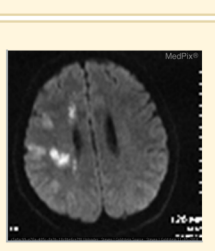

Reference 5: question = are regions of the brain infarcted?; answer = yes; 0.30 BGE question + 0.70 DINOv3 similarity = 0.832098.

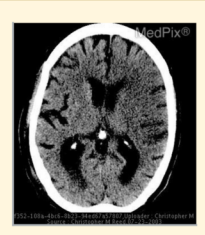

Reference 6: question = is there acute blood present on this head ct?; answer = no; 0.30 BGE question + 0.70 DINOv3 similarity = 0.828941.

### 3 User message: query image and required response schema

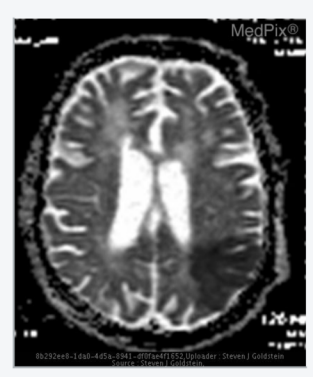

**is this consistent with an acute infarction?**

**Ground truth (evaluation only): yes**

```
Use the retrieved medical VQA examples as visual and question-answer
context. Answer the query image question conservatively from the
query image.

Question: is this consistent with an acute infarction?

Return only one compact JSON object with exactly this schema:
{"answer":"short answer","confidence":0.0,"evidence":"eight words
max"}. No markdown, no extra text.
```

### 4 Exact Gemma4 outputs

× **No-ICL**

{"answer": "no", "confidence": 0.95, "evidence": "ADC map shows high signal, no restricted diffusion"}

× **Fixed-6 ICL**

{"answer": "no", "confidence": 0.9, "evidence": "no restriction on ADC map"}

× **Random-6 ICL**

{"answer": "no", "confidence": 0.9, "evidence": "ADC map shows increased diffusion (bright signal)"}

✓ **RG-ICL (prompt shown)**

{"answer": "yes", "confidence": 0.9, "evidence": "hyperintense signal on diffusion weighted imaging"}

**Supplementary Fig. 6 | Exact prompt construction and case-level model outputs.** a, LAG classification case. The system message, six CLIP+DINOv3-retrieved image-text demonstrations, query image and response schema are reproduced verbatim in their model-input order. The displayed RG-ICL prompt produced the correct non-glaucoma JSON response, whereas No-ICL, Fixed-6 ICL and Random-6 ICL predicted glaucoma. b, VQA-RAD case. The complete BGE+DINOv3 RG-ICL prompt contains six image-question-answer demonstrations followed by the query ‘Is this consistent with an acute infarction?’ and the required JSON schema. RG-ICL returned the correct answer ‘yes’, whereas the three comparison conditions answered ‘no’.